\documentclass[a4paper]{article}[11pts]

\usepackage[english]{babel}
\usepackage[utf8x]{inputenc}
\usepackage[T1]{fontenc}

\normalsize

\usepackage[a4paper,top=1in,bottom=1in,left=1in,right=1in,marginparwidth=1in]{geometry}

\usepackage{setspace}
\usepackage{amsmath}
\usepackage{amssymb} 
\usepackage{amsthm}
\usepackage{amsfonts, mathtools}
\usepackage{algorithm,algpseudocode}
\usepackage{bm}
\usepackage{bbm}
\usepackage{comment}
\usepackage{graphicx}
\usepackage{multirow, booktabs}
\usepackage{caption}
\usepackage{subcaption}
\usepackage[colorinlistoftodos]{todonotes}
\usepackage[colorlinks=true, allcolors=blue]{hyperref}

\newcommand{\prob}{\mathbb{P}}

\usepackage{natbib}

\usepackage{natbib}

\title{Deep Learning for Choice Modeling}
\author{}
\date{}

\author{Zhongze Cai$^*$, Hanzhao Wang$^*$, Kalyan Talluri, Xiaocheng Li}
\date{\small 
Imperial College Business School, Imperial College London\\
(z.cai22, h.wang19, kalyan.talluri, xiaocheng.li)@imperial.ac.uk}

\begin{document}
\maketitle

\onehalfspacing

\let\thefootnote\relax\footnotetext{$^*$ Equal contribution.}

\begin{abstract}
Choice modeling has been a central topic in the study of individual preference or utility across many fields including economics, marketing, operations research, and psychology. While the vast majority of the  literature on choice models has been devoted to the analytical properties that lead to managerial and policy-making insights, the existing methods to learn a choice model from empirical data are often either computationally intractable or sample inefficient. In this paper, we develop deep learning-based choice models under two settings of choice modeling: (i) feature-free and (ii) feature-based. Our model captures both the intrinsic utility for each candidate choice and the effect that the assortment has on the choice probability. Synthetic and real data experiments demonstrate the proposed models’ performances in terms of the recovery of the existing choice models, sample complexity, assortment effect, architecture design, and model interpretation. 
\end{abstract}

\section{Introduction}

Choice models are used to explain or predict the choice behaviour of an individual or segment among a set of alternatives. It has been widely studied in various fields including economics, marketing, operations research, and psychology. In practice, choice models help the firm to understand customer behaviour and serve as pillar for revenue management to support decisions of assortment, marketing, and pricing. In this paper, we develop deep learning-based choice models with an emphasis on learning the choice model from data. Specifically, we consider two settings of choice modeling: (i) the feature-free setting where no additional feature is available and (ii) the feature-based setting to utilize the product and customer features.

\paragraph{Probabilistic choice models}\

There is a rich literature on choice modeling under the feature-free setting. The choice models aim to characterize the probability of choosing a product given an \textit{assortment} (a set of offered products). The prevalent choice models include multinomial logit model (MNL) , Markov chain choice model (MCCM) \citep{blanchet2016markov}, non-parametric choice model (NP) \citep{farias2009data}, mixture choice model \citep{mcfadden2000mixed}, etc. The existing literature focuses mainly on the analytical properties of these models and the corresponding revenue optimization problems. However, the learning/parameter estimation of these models remains less studied and the solutions so far have been unsatisfactory.

Existing methods to learn a choice model include:
\begin{itemize}
\item Maximum likelihood estimation (MLE): The method of MLE is the most standard approach to learn a probabilistic model. However, for most of the choice models, the likelihood function is often either non-convex or suffers from over-parameterization, with a few exceptions such as MNL and featured-based MNL.
\item Expectation maximization (EM): The EM algorithm can be used to learn choice models with latent variables such as MCCM and the mixture MNL model. There is generally no guarantee for the algorithm to converge to the global optimal. Besides, for the case of MCCM \citep{csimcsek2018expectation}, the algorithm needs to impute information related to the whole sample path (treated as unobserved variables) with merely the observation of the existing (final) state, which can result in a large variance. 
\item Linear programming (LP)-based method: The LP formulation is usually used to learn a non-parametric choice model \citep{farias2009data, chen2022decision}. Sparsity-promoting objective is often adopted to ensure statistical/sample efficiency. However, the LP-based approach is computationally unscalable in that the number of decision variables in LP usually scales exponentially with the number of underlying products. This prevents the approach to fit a choice model for more than a few tens of products.
\end{itemize}

Compared to the classic choice models, our deep learning-based choice models have two advantages. First, it has a large model capacity, and empirically, we show that one neural network model with a fixed architecture is capable of recovering both simple choice models such as MNL and complex choice models such as MCCM and NP. Second, it enables an effective learning procedure with better sample and computation efficiency than the above listed methods.

\paragraph{Neural network models in utility modeling}\

Another line of literature well utilizes the availability of the product and customer features. The works along this line usually model the latent utility of each product as a parameterized (neural network) function of the features and then relate the choice probabilities with the latent utilities through an MNL model. The learning of these models is essentially to recover the function parameters through the observations of customer's choice behaviour.

The idea is originated from \cite{bentz2000neural} which uses neural network to capture the non-linear effects of features on the latent utility. A few recent works \cite{wang2020deep,han2020neural,sifringer2020enhancing, wong2021reslogit,aouad2022representing} study different application contexts and explore different neural network architectures for the feature-utility mapping. In addition to the product and customer features, \cite{gabel2022product} also encode history purchase information. \cite{chen2021estimating} take a different route and consider the random forest model as the mapping function. 

We term this line of works as deep learning-based utility modeling instead of choice modeling, because these works mainly focus on predicting the unobserved product utilities with the observed features. However, unlike the classic literature on choice modeling, this line of works pays little attention to the \textit{assortment effect} -- the interactions between the products within (and possibly outside) the set of offered products. This brings two negative implications. First, some of these existing neural network models require all the training samples to have a fixed-size assortment. This requirement can be hardly met in many application contexts due to reasons such as product availability and profitability. Second, it fails to capture the effect that the assortment has on the choice probability (see Table \ref{tab:Newspaper_example} in the next section for a real-world example). Intuitively, a customer's choice behaviour is not only determined by product and customer features, but also affected by the assortment offered to the customer in a complicated manner. In fact, to model the assortment effect is one of the main theme of the feature-free choice modeling literature. 

Another common drawback of the existing deep learning approaches is that they all require the availability of some product or customer features. When there is no feature available, these models will degenerate and output a uniform choice probability. Differently, our paper proposes the first deep learning-based choice model to handle the feature-free setting. Also, we propose the first neural network models to explicitly capture the assortment's effect on choice probabilities and thus allow the training data to have variable assortment size. Importantly, we make a distinction between the feature-free model and the feature-based model not by the availability of the features, but by whether the downstream application is interested in a population-level or personalized choice model. 

To summarize, our contributions are as follows:
\begin{itemize}
\item We formulate the problem of learning a choice model from data as a generative modeling task and identify the challenges of the task. 
\item We develop neural network models for both the settings of feature-free and feature-based. 
\item We conduct extensive numerical experiments to illustrate the proposed models in terms of the recovery of the existing choice models, sample complexity, assortment effect, architecture design, and model interpretation.
\end{itemize}

\section{Problem Setup}

Consider a set of $n$ potential products $\mathcal{N}=\{1,2,...,n\}$ that can be offered to customers. A discrete choice model is a probabilistic model that describes the customers' choice behaviour under various \textit{assortments}. An assortment $\mathcal{S}\subset \mathcal{N}$ is a subset of $\mathcal{N}$ and it denotes the products offered to the customer to choose from. Practically, the assortment can be determined by the product availability (such as hotel rooms and flights), and the seller may also decide the assortment to maximize the profits (by removing less profitable products from the assortment). The choice model prescribes the probability of choosing product $i$ conditional on the assortment $\mathcal{S}$:
$$\prob(i|\mathcal{S}) \text{ for all } i\in\mathcal{N} \text{ and } \mathcal{S}\subset \mathcal{N}.$$
In particular, $\prob(i|\mathcal{S})=0$ for $i\notin \mathcal{S},$ i.e., the customer cannot choose a product not offered in the assortment. In this way, a choice model 
\begin{equation}
    \mathcal{M} = \{\prob(i|\mathcal{S}): \mathcal{S}\subset \mathcal{N}\}
    \label{choice_model}
\end{equation}
dictates $2^n-1$ probability distributions, each of which corresponds to one possible assortment. In practical contexts, there is usually a ``no-purchase'' option where the customer ends up with purchasing no product from the offered assortment. The no-purchase option can be captured by one product (say, $n$-th) in $\mathcal{N}.$ If no purchase is allowed, then we always have $n\in \mathcal{S}.$

A tabular parameterization of the choice model $\mathcal{M}$ generally requires $\Omega(n\cdot 2^n)$ parameters. Additional structures are introduced to facilitate efficient learning and inference. 

\subsection{Random Utility Model (RUM)}

The \textit{random utility model} (RUM) is the most prevalent class of choice models. It assigns a utility for each product and models the choice behavior according to the principle of utility maximization. Specifically, the utility of the $i$-th product 
$$U_i=u_i+\epsilon_i \text{ for } i\in\mathcal{N}.$$
Here $u_i$'s are deterministic and represent the mean utility of purchasing the $i$-th product among the population. $\epsilon_i$ are mean-zero, possibly dependent random noises that capture the utility heterogeneity among the customers. Under the RUM,
\begin{equation}
\prob(i|\mathcal{S}) \coloneqq \prob\left(U_{i} = \max_{j\in\mathcal{S}} U_{j}\right)
\label{eqn:RUM}
\end{equation}
For simplicity, we assume $\epsilon_i$'s are continuous random variables so that the utility maximizer is always unique. Different distributions of $\epsilon_i$'s specialize the general-form RUM into different concrete choice models. In this paper, we use the following two as running examples of RUM.

\textbf{Multinomial Logit Model (MNL).} The celebrated multinomial logit model assumes $\epsilon_i$'s take mean-zero Gumbel distribution (with scale parameter 1). With the Gumbel-max trick \cite{gumbel1954statistical}, the choice probability follows,
\begin{equation}
    \prob(i|\mathcal{S}) \coloneqq \frac{\exp(u_i)}{\sum_{j\in\mathcal{S}}\exp(u_{j})}.
    \label{eqn:MNL}
\end{equation}
When there is a feature vector $f_i$ associated with each product, then one can represent the deterministic utility $u_i=\theta^\top f_i$ and obtain the linear MNL model \cite{javanmard2020multi}. Also, several works replace $u_i$ with a neural network function of $f_i$ to obtain the DeepMNL model \citep{bentz2000neural,sifringer2020enhancing,wang2020deep}.

\textbf{Markov Chain Choice Model (MCCM).} The MCCM \cite{blanchet2016markov} defines a discrete-time Markov chain on the space of products $\mathcal{N}$ and models each customer's choice through the realization of a sample path. Specifically, it assumes a customer arrives to product $i$ with probability $\lambda_i$ (interpreted as the initial state of the Markov chain). Then if the product is in the assortment $\mathcal{S}$, the customer purchases it and leaves. Otherwise, the customer transitions from product $i$ to product $j$ with probability $\rho_{ij}$. Naturally, $\sum_{i\in\mathcal{N}} \lambda_i=1$ and $\sum_{j\in\mathcal{N}} \rho_{ij}=1$ for all $i\in\mathcal{N}.$ Let $X_t\in\mathcal{N}$ denote the product (state) that the customer visits at time $t$, and define the first hitting time $\tau=\{t\ge0: X_t\in\mathcal{S}\}.$ Then, under the MCCM,
$$\prob(i|\mathcal{S}) \coloneqq \prob(X_{\tau}=i).$$
The MCCM is proved as a generalization of MNL; with more number of parameters, the MCCM has a larger model capacity to capture more complex choice behavior.
 
Both MNL and MCCM can be viewed as a RUM. This fact is evident for MNL based on its definition, but no so much for MCCM at first sight. \cite{berbeglia2016discrete} first establishes MCCM as a RUM through a random walk argument. 

\subsection{Beyond RUM}
\label{sec:beyond_rum}
Under the RUM, although the choice probability is determined jointly by the utilities and the assortment, the contributions of these two aspects are separate. Specifically, the assortment will only affect the final choice through the constraint on the candidate products (to obtain the maximum utility in \eqref{eqn:RUM}), but it will not affect the utilities, neither $u_i$ nor $\epsilon_i$. In other words, RUM assumes that the utility $U_i$'s distribution are independent from and not affected by the assortment $\mathcal{S}.$ A probabilistic statement of this principle is known as the \textit{regularity} property \citep{gallego2019revenue}. 

However, as remarked by \cite{chen2022decision}, there is an increasing body of experimental evidence, arising in the various fields, which suggests that the aggregate choice behavior of customers is not always consistent with this regularity property and thus often violates the principle of the RUM. For instance, the following table summarizes a behavioral experiment involving subscriptions to The Economist (reproduced from \citep{ariely2008predictably}).

\begin{table}[ht!]
    \centering
    \begin{tabular}{cc|cc}
    \toprule
        &Option & Price & \# Subscribers  \\
    \midrule
    \multirow{2}{*}{Case I}   &Internet-Only  & \$59 &68 \\
        &Print-\&-Internet & \$125 & 32\\
    \bottomrule
    \end{tabular}
    \begin{tabular}{cc|cc}
    \toprule
        &Option & Price & \# Subscribers  \\
    \midrule
    \multirow{3}{*}{Case II}    &Internet-Only  & \$59 &  16  \\
        &Print-Only & \$125 & 0 \\
        &Print-\&-Internet & \$125 & 84 \\
    \bottomrule
    \end{tabular}
    \caption{Behavioral experiment on the subscriptions to The Economist \citep{ariely2008predictably}.}
    \label{tab:Newspaper_example}
\end{table}

In Case I, an assortment with two products (options) are available, while in Case II, a clearly inferior option of Print-Only is added. The addition of Print-Only twists customer utilities for the other two options, which is reflected by the change in the numbers of subscribers. Generally, it means that the assortment $\mathcal{S}$ will not only determine the products to choose from, but also affect the intrinsic utility of each product $U_i$'s. 

The practical example above motivates the study of non-parametric choice models such as \citep{farias2009data} and \citep{chen2022decision}. Also, it distinguishes our models with the existing  deep learning-based choice/utility models in that our models take into account this assortment effect when predicting the choice probabilities. In the following, we introduce the permutation-based non-parametric choice model as a representative that does not belong to the class of RUM.

\textbf{Non-parametric (NP) Choice Model.} The NP choice model  \cite{farias2009data, farias2013nonparametric} assumes that there exists a distribution $\lambda: \text{Perm}_\mathcal{N} \rightarrow[0,1]$ over the set $\text{Perm}_\mathcal{N}$ of all possible permutations of the products. There are $n!$ customer types and each customer type has a preference list of the products corresponding to one permutation in $\text{Perm}_\mathcal{N}$. Customers always purchase the most preferred product in the perference list. Then the choice probability is given by
$$\prob(i|\mathcal{S}) \coloneqq\sum_{\sigma \in \text{Perm}_i(\mathcal{S})} \lambda(\sigma) $$
where the set $$\text{Perm}_i(\mathcal{S})\coloneqq\{\sigma\in\text{Perm}_{\mathcal{N}}: \sigma(i)<\sigma(j)\text{ for all } j\neq i \in \mathcal{S}\}$$
contains all customer types/permutations under which product $i$ is the most preferable product in the assortment $\mathcal{S}.$

The NP choice model has exponentially many parameters and thus a larger model capacity than MNL and MCCM. There are also other choice models include mixed multinomial logit model \citep{mcfadden2000mixed}, nested logit model \citep{wen2001generalized}, tree-based choice model \citep{chen2022decision}, etc.


\subsection{An application of choice modeling}

An important application of choice modeling is the assortment optimization problem. The seller aims to find the assortment that maximizes the profits
\begin{equation}
\max_{\mathcal{S}} \sum_{i\in\mathcal{N}}r_{i} \prob(i|\mathcal{S})
\label{assort_opt}
\end{equation}
where $r_i$ is the profits/revenue of selling one unit of product $i$. For choice models such as MNL and MCCM, there exist polynomial-time algorithms to solve the problem. For general choice models, heuristic algorithms are proposed, such as the revenue-ordered strategy \citep{talluri2004revenue} and the greedy strategy \citep{jagabathula2014assortment}.

\subsection{Challenges of learning choice model from data}

The learning of a choice model refers to the identification of a choice model $\mathcal{M}$ from the available data
\begin{equation}
\mathcal{D}=\{(i_k, \mathcal{S}_k), k=1,...,m\}.\label{train_data}    
\end{equation}
Here each observation consists of a choice $i_k$ made under the assortment $\mathcal{S}_k.$ Throughout this paper, we use $n=|\mathcal{N}|$ to denote the number of products, and $m$ to denote the number of samples. We make the following two remarks.

First, the learning of a choice model should be viewed as a generative modeling task rather than discriminative modeling. Notably, exponentially many probability distributions -- one for each assortment as in \eqref{choice_model} -- need to be learned simultaneously from the data. This challenge can be further exaggerated when the choice model contains exponentially many parameters such as \citep{farias2009data, chen2022decision}. 

Second, the learning task suffers from the \textit{efficiency issue} in the rare-event probability estimation \citep{asmussen2007stochastic}. Consider learning a choice model with $n$ products. The probability of choosing each product is on the order of $1/n$ under a large assortment. Consequently, the standard deviation of the estimator is also on the order of $1/\sqrt{n}$ (taking Bernoulli variable as an example), which is significantly larger than the mean. Thus it inevitably requires a larger amount of samples to reduce the variance and accurately estimate such small probability.

\section{Feature-free Choice Modeling}

\begin{table*}[ht!]
    \centering
    \begin{tabular}{c|c|cc|cc|cc}
    \toprule
    &\multirow{2}{*}{\# Samples}&\multicolumn{2}{c|}{MNL}  &\multicolumn{2}{c|}{MCCM}&\multicolumn{2}{c}{NP}  \\
        &&$|\mathcal{N}|=20$&$|\mathcal{N}|=50$  &$|\mathcal{N}|=20$&$|\mathcal{N}|=50$ &$|\mathcal{N}|=20$&$|\mathcal{N}|=50$ \\ 
         \midrule
   Uniform & -- & 2.51 & 3.4 &2.52 & 3.44 & 2.50&3.43\\
\midrule
   \multirow{3}{*}{MNL-MLE}  & 1,000&	1.87&2.65&1.96&0.50&1.73&2.50\\
   & 5,000&1.86& 2.63& 1.96& 0.50& 1.71& 2.48\\
   &100,000&1.86& 2.62&1.95& 0.52& 1.71& 2.47\\
 \midrule
  \multirow{3}{*}{MCCM-EM}  & 1,000&1.86&2.67&1.92&0.54&1.67&2.35\\
   &5,000&1.86 &2.66 &1.88 &0.53 &1.60 &2.35 \\
   & 100,000 &1.87 &2.66	&	1.88&0.51	&	1.62&2.37	\\
\midrule
   \multirow{3}{*}{Gated-Assort-Net}  & 1,000&2.00&3.04& 1.97& 0.54& 1.69&2.71\\
   & 5,000& 1.89& 2.81& 1.88&0.48&1.55&2.43\\
   &100,000& 1.86& 2.63& 1.84& 0.40&1.50&2.23\\
\midrule
   \multirow{3}{*}{Res-Assort-Net}  & 1,000&2.05&3.05& 2.03& 0.61&1.80&2.79\\
   & 5,000&1.93&2.82&1.91&0.56&1.56& 2.51\\
   &100,000& 1.86& 2.64& 1.84&0.41&1.49& 2.22\\
\midrule
  Oracle & -- & 1.86&	2.62&	1.82&	0.38&	1.42&	2.13\\
   \bottomrule
\end{tabular}
\caption{Feature-free choice modeling: The training data are generated from the model of MNL, MCCM, and NP with randomly generated true parameters. The number of samples refers to the number of training samples, and additional 5,000 samples are reserved to validate the training procedure. The reported numbers are the average out-of-sample cross-entropy (CE) loss over 10,000 test samples and 10 random trials. The row Uniform denotes the performance of a uniform distribution. MNL-MLE implements the standard maximum likelihood estimation to estimate a MNL model, while MCCM-EM implements the expectation maximization algorithm to fit a MCCM model. GAsN has one hidden fully-connected layer, and RAsN has two residual blocks. The row Oracle denotes the CE loss of the true model.}
    \label{tabRP}
\end{table*}

We first discuss the plain setting where there is no feature associated with each product or each customer, and the training data is $\mathcal{D}$ (as in \eqref{train_data}). This is the most straightforward and widely studied setting in the literature of choice modeling. We use this setting to illustrate the prototypes of our two neural network models and will incorporate the product and customer features in the later sections.

\begin{figure}[ht!]
    \centering
    \begin{subfigure}[b]{0.24 \textwidth}
        \centering
        \includegraphics[height=2.7in]{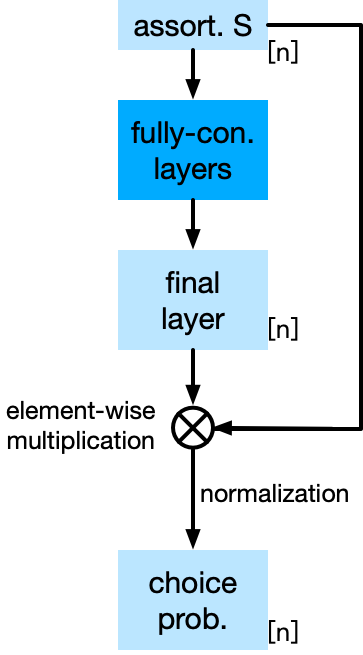}
        \caption{Gated-Assort-Net}
    \end{subfigure}%
    \begin{subfigure}[b]{0.24 \textwidth}
        \centering
        \includegraphics[height=2.7 in]{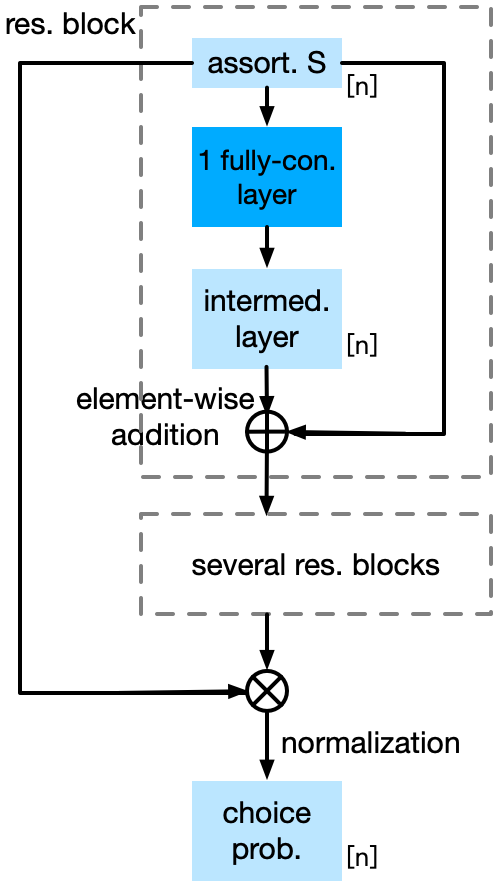}
        \caption{Res-Assort-Net}
    \end{subfigure}
    \caption{Feature-free choice networks. $[n]$ denotes the dimension of the corresponding later.}
    \label{fig:net_no_feat}
\end{figure}

 Mathematically, a neural network maps the assortment input $S \in \{0,1\}^n$ to the output $Y\in [0,1]^n$,
$$Y=g_\theta(S)$$
where $\theta$ encapsulates all the model parameters. Here $S=(s_1,...,s_n)$ is a binary encoding of an assortment $\mathcal{S}$ where $s_i=1$ if $i\in\mathcal{S}$ and $s_i=0$ if $i\notin\mathcal{S}$. The output $Y=(Y_1,...,Y_n)$ is a vector supported on the probability simplex with $Y_i=\prob(i|\mathcal{S})$. 

Figure \ref{fig:net_no_feat} describes our two neural network models:

Gated-Assort-Net (GAsN): It takes the assortment vector $S$ as its input layer and runs through a number of fully connected layers. Finally, it uses the assortment $S$ again to create an output gate to ensure $\prob(i|\mathcal{S})=0$ if $i\notin \mathcal{S}$.

Res-Assort-Net (RAsN): It implements the idea of residual learning \citep{he2016deep} and involves the input assortment $S$ recursively throughout the propagation. As in the Gated-Assort-Net, it uses the assortment to create an output gate to ensure $\prob(i|\mathcal{S})=0$ if $i\notin \mathcal{S}$.

Throughout the paper, we use the cross-entropy (CE) loss to train all our neural network models.

Table \ref{tabRP} summarizes our first group of experiments. We compare the performance of GAsN and RAsN against several benchmark methods of learning choice models, as well as the oracle performance of the true distribution (equivalently, the negative entropy of the true distribution). The training data are generated from the true models of MNL, MCCM, and NP. We make the following observations. First, both GAsN and RAsN benefit from a larger sample size, while the performances of maximum likelihood estimation (MLE) and expectation maximization (EM) do not improve as the sample size increases. Second, when the true model is MNL, we know the method of MLE is provably asymptotically optimal and this is unsurprisingly verified from our experiment. But when the true model becomes more complex, such as MCCM and NP, the neural network models show their advantage. Third, it requires a large sample size to learn a complex true model. For example, under the true model of MCCM or NP, none of these methods give a satisfactory performance when the sample size is $1,000$. Overall, GAsN and RAsN perform consistently well in recovering these true models with sufficiently amount of data such as $m=5000$. They have the model capacity to capture complex structures and they are also capable of fitting a simpler true model such as MNL. In comparison, the existing choice models either suffer from too little model capacity such as MNL, or have no reliable method for parameter estimation such as MCCM or NP. It is widely acknowledged that the models of MCCM and NP enjoy good analytical properties, but without an effective learning/estimation algorithm, these models can only provide qualitative insights but not reliable decision support.

Two other important aspects are (i) the assortment distribution in the training data and (ii) the depth and width of the two network models. For (i), we find the performance of the neural network model is quite robust in terms of the generalization on the assortment domain. Specifically, we train the neural network model with one distribution to generate the assortments $\mathcal{S}_k$'s (in training data $\mathcal{D}$) and find that it performs well on the test data where the assortments are generated from a different assortment distribution (also known as the out-of-domain performance). For (ii), we vary the depth and width of the neural network models and test their performances under the settings of (i) feature-free and (ii) with features. The default design of GAsN and RAsN has the intermediate layers of the same dimension $n$ as the input layer. We find that to fit a moderate size choice model with 20-50 products, a depth of 1-2 hidden layers for GAsN or 2-3 residual blocks for RAsN is sufficient. And also, the performance of the model remains stable with wider width for the intermediate layers. We defer the details of these two aspects to Appendix A.1 and A.2 respectively.

\section{Feature-based Choice Modeling}

\begin{figure}[ht!]
    \centering
    \includegraphics[scale=0.25]{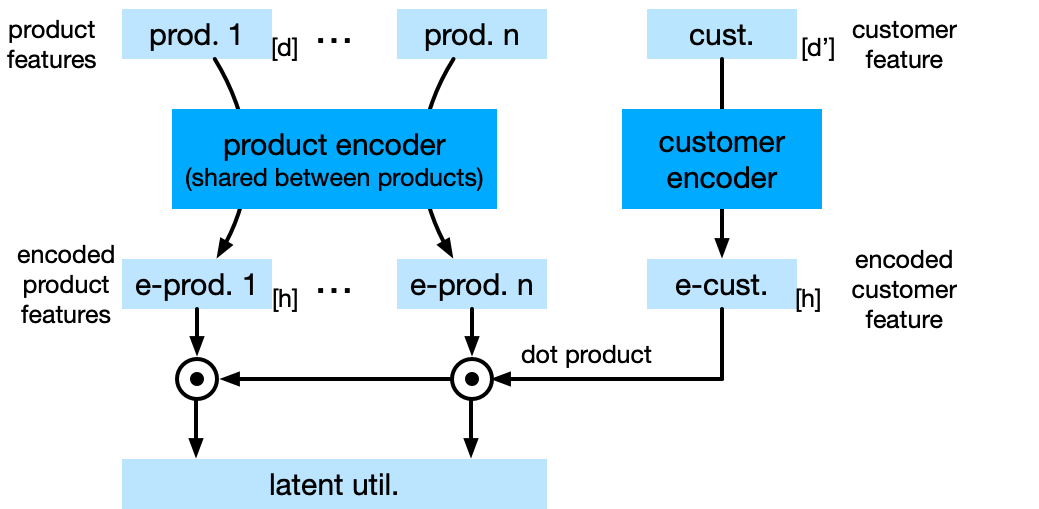}
    \caption{Feature encoder: Product features and customer features are encoded and then taken inner product to obtain the latent utility for each product. $d$ is the dimension of the product features, and $d'$ is the dimension of the customer features. Both features are encoded to $h$-dimensional latent features, and the latent utilities are obtained by the inner product of the corresponding latent features.}
    \label{fig:feat_encode}
\end{figure}

Now we extend our neural network models to the setting of choice modeling with features. We first make a distinction between product feature and customer feature:

\begin{itemize}
\item Product feature: there is a feature vector $f_i\in\mathbb{R}^d$ associated with each product $i\in\mathcal{N}$. We also refer to these features as \text{static} features as they usually remain the unchanged over all the samples in the training data $\mathcal{D}$. 
\item Customer feature: Sample in $\mathcal{D}$ represents different customers, each of which is associated a feature vector $g_k\in\mathbb{R}^{d'}$. With these features, the dataset is augmented as
$$\mathcal{D} = \{(i_k, \mathcal{S}_k, g_k), k=1,...,m\}.$$ 
We refer to these features as \textit{dynamic} features as they may vary across different samples. Modeling-wise, if there are product features that change over time, we can simply see them as dynamic features.
\end{itemize}

\begin{figure}[ht!]
    \centering
    \begin{subfigure}[b]{0.25\textwidth}
        \centering
        \includegraphics[height=3.5in]{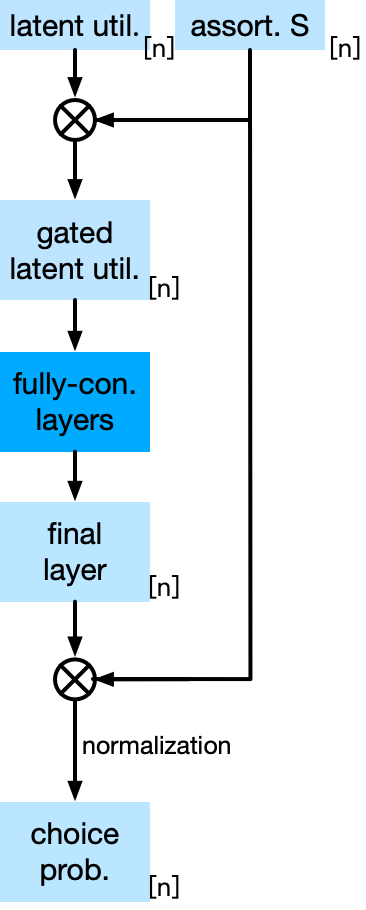}
        \caption{Gated-Assort-Net(f)}
    \end{subfigure}%
    \begin{subfigure}[b]{0.25\textwidth}
        \centering
        \includegraphics[height=3.5in]{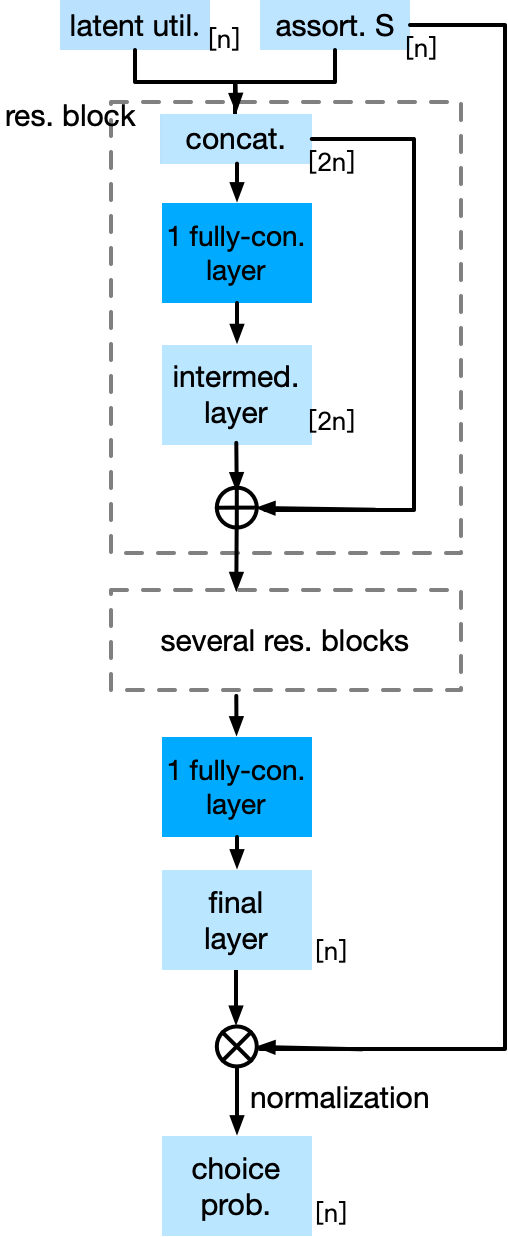}
        \caption{Res-Assort-Net(f)}
    \end{subfigure}
    \caption{Choice networks with features: The networks take (i) the latent utilities from feature encoder (in Figure \ref{fig:feat_encode}) and (ii) the assortment vector as inputs.}
    \label{fig:net_feat}
\end{figure}

Figure \ref{fig:feat_encode} and Figure \ref{fig:net_feat} describe our choice networks with features. The neural networks inherit the architectures in the feature-free setting, but both take an additional vector of inputs which we call as the latent utilities. Both feature-based networks of GAsN(f) and RAsN(f) encode product features and customer features to obtain one latent utility for each product. The product encoder is an $l$-layer fully-connected network shared by all the products ($l=1$ in our experiments). By default, the number of nodes in the intermediate layers of the encoder is the same as the dimension of the encoder's input. When there is no available customer feature, we can simply treat the customer feature as 1 for all the samples.

\subsection{Unnecessity of static features}

\begin{table}[ht!]
\centering
\begin{tabular}{c|cc}
\midrule
  &  MNL(f)  & MCCM(f) \\
\midrule
 MNL(f)-MLE & 2.903 & 2.212\\
 \midrule
Gated-Assort-Net &  2.921& 2.010\\
Gated-Assort-Net(f) & 2.918 & 2.010\\
\midrule
Res-Assort-Net & 2.914 & 1.998\\
Res-Assort-Net(f) & 2.914 & 1.985\\
\midrule
Oracle & 2.900 & 1.932\\
\bottomrule
\end{tabular}
\caption{Choice modeling with only product (static) features: The training data (with $n=50$, $m=100,000$, and $d=5$) are generated from the feature-based version of MNL and MCCM (see Appendix B.1 for details). The reported numbers are the average out-of-sample cross-entropy (CE) loss over 10 random trials. The benchmark method implements MLE for the featured-based MNL model.}
\label{tabFeat1}
\end{table}

Table \ref{tabFeat1} presents a synthetic experiment with only product (static) features to illustrate a new insight for choice modeling with features. The experiment first ignores all the product features and trains GAsN and RAsN as in the feature-free setting. Then it takes into account the product features and trains the feature-based version of the two neural networks. We find that both versions of the networks are capable of recovering the true model from the comparison against the oracle performance, and that the two versions of each neural network achieve a similar performance. This might be counter-intuitive at the first sight, because the product features may contain useful information about the product's utility, and ignoring the features may accordingly hurt the performance. However, we argue that the product utilities are essentially encoded in the choice data $(i_k, \mathcal{S}_k)$'s but not in the product features. In fact, the experiment shows that such utilities can be learned by the neural networks of GAsN and RAsN without explicitly accessing the product features. An alternative interpretation is that the product utility here should be viewed as a population-level utility that reflects the preference or popularity of a product over the whole population. Things become different when customer features are also available. In that case, the choice behaviour of each data sample is determined by the personalized utilities, and thus the product and customer features become indispensable.

We emphasize that whether to include the (dynamic) customer features in choice modeling is determined not by the availability of the features, but by the downstream application. Recall that the assortment optimization problem \eqref{assort_opt} aims to find an assortment that maximizes profits under a choice model. For a brick-and-mortar store or an online retailer where the personalized assortment is not allowed, customer features should not be used to fit the choice model even if they are available, because one is more interested in the population-level choice behavior. This explains why the majority of the literature on choice modeling has been devoted to feature-free models. And it also underscores the potential wide applicability of feature-free networks of GAsN and RAsN despite their simple architectures. 

\section{Experiments on Real Data}

In this section, we conduct numerical experiments on two real datasets -- SwissMetro (\url{https://transp-or.epfl.ch/pythonbiogeme/examples_swissmetro.html})  and Expedia Search (\url{https://www.kaggle.com/competitions/expedia-personalized-sort/overview}). These two datasets contain dynamic customer/product features; that is, the features associated with each training samples may be different with each other. The SwissMetro dataset is a public survey data set to analyze traveller preference among three transportation modes, and the Expedia dataset consists of hotel search records and the associated customer choices. We refer more details of these two datasets to Appendix \ref{sec: real_data_exp}.

\begin{table}[ht!]
    \centering
    \begin{tabular}{c|c|c|c|c}
    \toprule
    &\multicolumn{2}{c|}{SwissMetro}  &\multicolumn{2}{c}{Expedia}\\
    & CE Loss & Acc.&CE Loss  & Acc. \\
    \midrule
    MNL-MLE&0.883&0.581&2.827& 0.298\\
MNL(f)-MLE&0.810&0.621&2.407& 0.331\\
\midrule
TasteNet
&0.681&0.698&2.436&0.332\\
DeepMNL
&0.741&0.655&2.374&0.344\\
Random Forest
&0.633&0.735&2.458&0.317\\
\midrule
Gated-Assort-Net(f)
&\textbf{0.598}&\textbf{0.759}&2.403&0.332\\
Res-Assort-Net(f)&
0.607&0.738&\textbf{2.325}&\textbf{0.355}\\
   \bottomrule
    \end{tabular}
    \caption{Performance on the SwissMetro and Expedia.}
    \label{tabExp}
\end{table}

We implement a number of benchmark models: MNL, feature-based MNL, TasteNet \citep{han2020neural}, DeepMNL \citep{wang2020deep, sifringer2020enhancing}, random forest \citep{chen2021estimating}. The MNL and feature-based MNL are learned by MLE. For both TasteNet and DeepMNL, it models the product utility $u_i$ as a function of the product and customer features and maps the utility to choice probability via the model of MNL \eqref{eqn:MNL}. Tastenet assumes the utility $u_i$ as a linear function of product features and uses the customer feature to determine the coefficients of the linear function, while DeepMNL concatenates the customer and product features and feeds both into a fully-connected network to obtain the utility $u_i$. Both models do not utilize the assortment other than the MNL part.

From the results in Table \ref{tabExp}, our two neural networks give a better performance than the benchmark models. In fact, our neural network models take the most natural architectures, and the key difference is that our models explicitly take the assortment as an input. This experiment result shows that even with the presence of product and customer features, the assortment can help to better explain customer choices. Comparing the two datasets, the Gated-Assort-Net(f) performs better on SwissMetro, while Res-Assort-Net(f) performs better on Expedia. Recall that the Res-Assort-Net(f) has a stronger usage of the assortment vector throughout the architecture than Gated-Assort-Net(f). Accordingly, one explanation can be that for the transportation setting, customer choices are less affected by the available options but more by their personal preference, but for the hotel search, the customer choices are more affected by the provided assortment which gives Res-Assort-Net(f) more advantage. As for the random forest model, it is trained as a discriminative model using both features and assortment as input, so we believe it also better potential with some further model recalibration.

\begin{figure}
    \centering
    \includegraphics[scale=0.4]{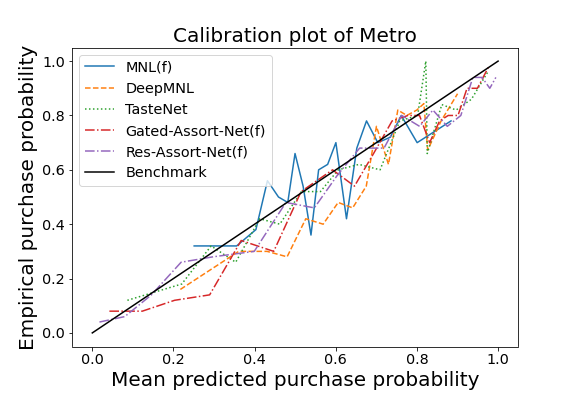}
    \caption{Overconfidence of predicting Metro. The predicted probabilities of the test samples are binned into $20$ bins with the same number of samples. For each bin we compute the mean predicted prob. as x-axis and the mean empirical prob. as y-axis. The benchmark is black solid line where $x=y$.}
    \label{fig:calibrate_swiss}
\end{figure}

\textbf{Model Calibration.} Another important aspect of probabilistic model is model calibration. Ideally, we hope the predicted probability matches the empirical/true probability. For deep learning models, a high accuracy may often come with an overconfidence in predictions (overestimation of the true probabilities) \cite{guo2017calibration}, which is called \textit{miscalibration}. Figure \ref{fig:calibrate_swiss} is the calibration plot of predicting Metro for the SwissMetro dataset, and we observe that all the neural network models stay mostly below the 45-degree reference line, i.e., suffer from an over-confidence in predicting the probability. The result points to more future study on other proper metric than CE loss and accuracy for evaluating choice model learning and effective methods for recalibrating choice models. We defer more visualizations of to Appendix C.


\section{Discussion and Future Directions}

The existing literature of choice modeling on the feature-free setting and that on the feature-based setting have been largely separate with each other. In this paper, we propose a unified deep learning-based framework that applies to both settings through a binary representation of the assortment and feature encoders. We believe there is still large room for performance improvement by further tuning the neural network architectures. We summarize our findings into the following three points. First, the deep learning-based choice models are capable of recovering existing choice models with an effective learning procedure. They become particularly effective when the underlying model/training data is too complex to be described by a simple model such as MNL and when there is a sufficient amount of training data (such as 5,000 samples for 20-50 products). Second, the modeling of assortment effect has always been a key component of choice models, which can be verified by both empirical evidence such as Table \ref{tab:Newspaper_example} and all the numerical experiments in this paper. Third, whether to include features in choice models is determined by the downstream application rather than feature availability, and feature-free choice models should be used when one is interested in understanding population-level choice behaviour. 

We conclude with the following two future directions. 

\textbf{Multiple-purchase choice model.} The existing choice models focus exclusively on modeling the single-choice behavior where only one product is chosen from the offered assortment. The multiple-purchase choice model is usually studied under an inverse optimization framework in the literature of revealed preference \citep{zadimoghaddam2012efficiently,amin2015online,birge2022learning}. The deep learning-based choice models provide a natural framework to study the multiple-choice behavior, and they can fit to such training data by twisting the loss function.

\textbf{Pricing and assortment optimization.} In a retailing context, the product price is another important factor that affects customer choice. Our current models did not account for this factor. One option is to include the price vector as another input to the neural network models. We believe it deserves more future works to study the associated price and assortment optimization problems.

\subsection*{Acknowledgment}
The authors thank Antoine Desir for helpful discussions and pointing to the two datasets for numerical experiments.

\bibliographystyle{informs2014}
\bibliography{main.bib} 

\begin{thebibliography}{33}
\expandafter\ifx\csname natexlab\endcsname\relax\def\natexlab#1{#1}\fi
\expandafter\ifx\csname url\endcsname\relax
  \def\url#1{{\tt #1}}\fi
\expandafter\ifx\csname urlprefix\endcsname\relax\def\urlprefix{URL }\fi
\expandafter\ifx\csname urlstyle\endcsname\relax
  \expandafter\ifx\csname doi\endcsname\relax
  \def\doi#1{doi:\discretionary{}{}{}#1}\fi \else
  \expandafter\ifx\csname doi\endcsname\relax
  \def\doi{doi:\discretionary{}{}{}\begingroup \urlstyle{rm}\Url}\fi \fi

\bibitem[{Amin et~al.(2015)Amin, Cummings, Dworkin, Kearns, and
  Roth}]{amin2015online}
Amin, Kareem, Rachel Cummings, Lili Dworkin, Michael Kearns, Aaron Roth. 2015.
\newblock Online learning and profit maximization from revealed preferences.
\newblock {\it Proceedings of the AAAI Conference on Artificial
  Intelligence\/}, vol.~29.

\bibitem[{Aouad and D{\'e}sir(2022)}]{aouad2022representing}
Aouad, Ali, Antoine D{\'e}sir. 2022.
\newblock Representing random utility choice models with neural networks.
\newblock {\it arXiv preprint arXiv:2207.12877\/} .

\bibitem[{Ariely and Jones(2008)}]{ariely2008predictably}
Ariely, Dan, Simon Jones. 2008.
\newblock {\it Predictably irrational\/}.
\newblock HarperCollins New York.

\bibitem[{Asmussen and Glynn(2007)}]{asmussen2007stochastic}
Asmussen, S{\o}ren, Peter~W Glynn. 2007.
\newblock {\it Stochastic simulation: algorithms and analysis\/}, vol.~57.
\newblock Springer.

\bibitem[{Barratt and Boyd(2022)}]{article}
Barratt, Shane, Stephen Boyd. 2022.
\newblock Fitting feature-dependent markov chains.
\newblock {\it Journal of Global Optimization\/}
  1--18\doi{10.1007/s10898-022-01198-0}.

\bibitem[{Bentz and Merunka(2000)}]{bentz2000neural}
Bentz, Yves, Dwight Merunka. 2000.
\newblock Neural networks and the multinomial logit for brand choice modelling:
  a hybrid approach.
\newblock {\it Journal of Forecasting\/} {\bf 19}(3) 177--200.

\bibitem[{Berbeglia(2016)}]{berbeglia2016discrete}
Berbeglia, Gerardo. 2016.
\newblock Discrete choice models based on random walks.
\newblock {\it Operations Research Letters\/} {\bf 44}(2) 234--237.

\bibitem[{Bierlaire(2018)}]{SwissMetro}
Bierlaire, M. 2018 \urlprefix\url{http://transp-or.epfl.ch/documents/
  technicalReports/CS_SwissmetroDescription.pdf}.

\bibitem[{Birge et~al.(2022)Birge, Li, and Sun}]{birge2022learning}
Birge, John~R, Xiaocheng Li, Chunlin Sun. 2022.
\newblock Learning from stochastically revealed preference.
\newblock {\it arXiv preprint arXiv:2206.01484\/} .

\bibitem[{Blanchet et~al.(2016)Blanchet, Gallego, and
  Goyal}]{blanchet2016markov}
Blanchet, Jose, Guillermo Gallego, Vineet Goyal. 2016.
\newblock A markov chain approximation to choice modeling.
\newblock {\it Operations Research\/} {\bf 64}(4) 886--905.

\bibitem[{Bodea et~al.(2009)Bodea, Ferguson, and Garrow}]{bodea2009data}
Bodea, Tudor, Mark Ferguson, Laurie Garrow. 2009.
\newblock Data set—choice-based revenue management: Data from a major hotel
  chain.
\newblock {\it Manufacturing \& Service Operations Management\/} {\bf 11}(2)
  356--361.

\bibitem[{Chen et~al.(2021)Chen, Gallego, and Tang}]{chen2021estimating}
Chen, Ningyuan, Guillermo Gallego, Zhuodong Tang. 2021.
\newblock Estimating discrete choice models with random forests.
\newblock {\it INFORMS International Conference on Service Science\/}.
  Springer, 184--196.

\bibitem[{Chen and Mi{\v{s}}i{\'c}(2022)}]{chen2022decision}
Chen, Yi-Chun, Velibor~V Mi{\v{s}}i{\'c}. 2022.
\newblock Decision forest: A nonparametric approach to modeling irrational
  choice.
\newblock {\it Management Science\/} .

\bibitem[{Farias et~al.(2009)Farias, Jagabathula, and Shah}]{farias2009data}
Farias, Vivek, Srikanth Jagabathula, Devavrat Shah. 2009.
\newblock A data-driven approach to modeling choice.
\newblock {\it Advances in Neural Information Processing Systems\/} {\bf 22}.

\bibitem[{Farias et~al.(2013)Farias, Jagabathula, and
  Shah}]{farias2013nonparametric}
Farias, Vivek~F, Srikanth Jagabathula, Devavrat Shah. 2013.
\newblock A nonparametric approach to modeling choice with limited data.
\newblock {\it Management science\/} {\bf 59}(2) 305--322.

\bibitem[{Gabel and Timoshenko(2022)}]{gabel2022product}
Gabel, Sebastian, Artem Timoshenko. 2022.
\newblock Product choice with large assortments: A scalable deep-learning
  model.
\newblock {\it Management Science\/} {\bf 68}(3) 1808--1827.

\bibitem[{Gallego and Topaloglu(2019)}]{gallego2019revenue}
Gallego, Guillermo, Huseyin Topaloglu. 2019.
\newblock {\it Revenue management and pricing analytics\/}, vol. 209.
\newblock Springer.

\bibitem[{Gumbel(1954)}]{gumbel1954statistical}
Gumbel, Emil~Julius. 1954.
\newblock {\it Statistical theory of extreme values and some practical
  applications: a series of lectures\/}, vol.~33.
\newblock US Government Printing Office.

\bibitem[{Guo et~al.(2017)Guo, Pleiss, Sun, and
  Weinberger}]{guo2017calibration}
Guo, Chuan, Geoff Pleiss, Yu~Sun, Kilian~Q Weinberger. 2017.
\newblock On calibration of modern neural networks.
\newblock {\it International conference on machine learning\/}. PMLR,
  1321--1330.

\bibitem[{Gupta and Hsu(2020)}]{gupta2020parameter}
Gupta, Arushi, Daniel Hsu. 2020.
\newblock Parameter identification in markov chain choice models.
\newblock {\it Theoretical Computer Science\/} {\bf 808} 99--107.

\bibitem[{Han et~al.(2020)Han, Zegras, Pereira, and Ben-Akiva}]{han2020neural}
Han, Yafei, Christopher Zegras, Francisco~Camara Pereira, Moshe Ben-Akiva.
  2020.
\newblock A neural-embedded choice model: Tastenet-mnl modeling taste
  heterogeneity with flexibility and interpretability.
\newblock {\it arXiv preprint arXiv:2002.00922\/} .

\bibitem[{He et~al.(2016)He, Zhang, Ren, and Sun}]{he2016deep}
He, Kaiming, Xiangyu Zhang, Shaoqing Ren, Jian Sun. 2016.
\newblock Deep residual learning for image recognition.
\newblock {\it Proceedings of the IEEE conference on computer vision and
  pattern recognition\/}. 770--778.

\bibitem[{Jagabathula(2014)}]{jagabathula2014assortment}
Jagabathula, Srikanth. 2014.
\newblock Assortment optimization under general choice.
\newblock {\it Available at SSRN 2512831\/} .

\bibitem[{Javanmard et~al.(2020)Javanmard, Nazerzadeh, and
  Shao}]{javanmard2020multi}
Javanmard, Adel, Hamid Nazerzadeh, Simeng Shao. 2020.
\newblock Multi-product dynamic pricing in high-dimensions with heterogeneous
  price sensitivity.
\newblock {\it 2020 IEEE International Symposium on Information Theory
  (ISIT)\/}. IEEE, 2652--2657.

\bibitem[{McFadden and Train(2000)}]{mcfadden2000mixed}
McFadden, Daniel, Kenneth Train. 2000.
\newblock Mixed mnl models for discrete response.
\newblock {\it Journal of applied Econometrics\/} {\bf 15}(5) 447--470.

\bibitem[{Nixon et~al.(2019)Nixon, Dusenberry, Zhang, Jerfel, and
  Tran}]{nixon2019measuring}
Nixon, Jeremy, Michael~W Dusenberry, Linchuan Zhang, Ghassen Jerfel, Dustin
  Tran. 2019.
\newblock Measuring calibration in deep learning.
\newblock {\it CVPR Workshops\/}, vol.~2.

\bibitem[{Sifringer et~al.(2020)Sifringer, Lurkin, and
  Alahi}]{sifringer2020enhancing}
Sifringer, Brian, Virginie Lurkin, Alexandre Alahi. 2020.
\newblock Enhancing discrete choice models with representation learning.
\newblock {\it Transportation Research Part B: Methodological\/} {\bf 140}
  236--261.

\bibitem[{{\c{S}}im{\c{s}}ek and Topaloglu(2018)}]{csimcsek2018expectation}
{\c{S}}im{\c{s}}ek, A~Serdar, Huseyin Topaloglu. 2018.
\newblock An expectation-maximization algorithm to estimate the parameters of
  the markov chain choice model.
\newblock {\it Operations Research\/} {\bf 66}(3) 748--760.

\bibitem[{Talluri and Van~Ryzin(2004)}]{talluri2004revenue}
Talluri, Kalyan, Garrett Van~Ryzin. 2004.
\newblock Revenue management under a general discrete choice model of consumer
  behavior.
\newblock {\it Management Science\/} {\bf 50}(1) 15--33.

\bibitem[{Wang et~al.(2020)Wang, Mo, and Zhao}]{wang2020deep}
Wang, Shenhao, Baichuan Mo, Jinhua Zhao. 2020.
\newblock Deep neural networks for choice analysis: Architecture design with
  alternative-specific utility functions.
\newblock {\it Transportation Research Part C: Emerging Technologies\/} {\bf
  112} 234--251.

\bibitem[{Wen and Koppelman(2001)}]{wen2001generalized}
Wen, Chieh-Hua, Frank~S Koppelman. 2001.
\newblock The generalized nested logit model.
\newblock {\it Transportation Research Part B: Methodological\/} {\bf 35}(7)
  627--641.

\bibitem[{Wong and Farooq(2021)}]{wong2021reslogit}
Wong, Melvin, Bilal Farooq. 2021.
\newblock Reslogit: A residual neural network logit model for data-driven
  choice modelling.
\newblock {\it Transportation Research Part C: Emerging Technologies\/} {\bf
  126} 103050.

\bibitem[{Zadimoghaddam and Roth(2012)}]{zadimoghaddam2012efficiently}
Zadimoghaddam, Morteza, Aaron Roth. 2012.
\newblock Efficiently learning from revealed preference.
\newblock {\it International Workshop on Internet and Network Economics\/}.
  Springer, 114--127.

\end{thebibliography}

\appendix

\section{More Experiments on Robustness and Generalization}

\subsection{Out-of-domain performance under different assortment distribution}
\label{sec:apx_ass_eff_A1}

In this experiment, we demonstrate whether a different distribution of the assortment  between training and testing will affect model performance, also known as the out-of-domain generalization. Specifically, we use one distribution to generate the assortment in the training data and test the performance with assortment generated from another distribution. Intuitively, this shows whether the model is capable of generalizing over the domain of assortment.

In the experiment, we generate four groups of training data:
\begin{itemize}
\item For the first dataset D-1, assortments are generated by first uniformly randomly choosing $s=|\mathcal{S}|$,  and then uniformly randomly picking $s$ products from $\mathcal{N}$ and form the assortment $S$.
\item For the second dataset D-2, assortments are generated from Bernoulli distribution. That is, for each product, with probability $1/2$ it appears in the assortment independently with others.
\item For the third dataset D-3, the assortment is given by first randomly specifying either the first half (product 1 to product $n/2$, suppose $n/2$ is integer for convenience) or the second half (product $n/2+1$ to product $n$) of the total products will definitely not appear in the assortment. Then we generate the assortment within the rest half following the same rule as in D-1. 
\item For the last dataset D-4, we require all the assortments to have a size of  $n/3$ or $n/3+1$ (again assume they are integers for convenience) products, and then pick products randomly into the assortment. This design is inspired by the results of \cite{gupta2020parameter}. 
\end{itemize}

The total number of products $n=30$ and each training dataset consists of $m=100,000$ samples. The underlying true model for generating the data is the MCCM model. 

\begin{table}[ht!]
\centering
\begin{tabular}{c|cccc}
\toprule
  & D-1  & D-2  & D-3 & D-4 \\
\midrule
D-1 & 2.364 & 2.603 & 1.820 & 2.298
\\
D-2 & 2.366 & 2.600 & 1.819 & 2.298
\\
D-3 & 2.636	& 2.781 & 1.819 & 2.383
\\
D-4 & 2.434 & 2.619 & 1.829 & 2.297
\\ \midrule
Mix & 2.366 & 2.604 & 1.819 & 2.299 \\
\midrule
Oracle & 2.358 & 2.595 & 1.817 & 2.293
\\
\bottomrule
\end{tabular}
\caption{Out-of-sample CE loss for out-of-domain performance of Gated-Assort-Net. Row headers denote the training dataset and column headers denote the testing dataset. The ``Mix'' header denotes a randomly mixed training set with equal proportion from the $4$ datasets, and the ``Oracle'' header denotes the performance of the true model.}
\label{table:AssortEffect}
\end{table}

We train a neural network model with each of the training data and test its performance on the other datasets. In addition, we train a neural network model with a mixture of training data from the four datasets. Table \ref{table:AssortEffect} summarizes the out-of-sample CE loss for each training-test pair. For illustration, we train a 1-layer Gated-Assort-Net. The row headers denote the training dataset , and the column headers denote the testing dataset. 

The result shows that, except for the dataset D-3, Gated-Assort-Net performs well in out-of-domain datasets. It is understandable that the datasets D-1 and D-2 are informative enough for a neural network model to catch the whole picture of the true underlying model. The dataset D-4 also enjoys this privilege, which is a bit surprising but also reinforces the findings in \cite{gupta2020parameter} (though not in an exact same spirit). The distribution of D-3 clearly fails to capture the interplay between the first half and the second half of the total products, and thus can not reach the performance of the oracle or generalize to different assortment distributions.

\textbf{Details on the true model generation:} 

The training data is generated from an MCCM. The arriving probability $\lambda_i$ for product $i$ is generated by:
$$
\lambda_i = \dfrac{\exp(\mu_i)}{\sum_{j=1}^{n} \exp(\mu_j)},
$$
where $\mu_i$, $i=1,...,n$ is identically and independently (i.i.d.) sampled by standard Normal distribution $\mathcal{N}(0,1)$. And the transition matrix $\rho$ is generated row by row. For all the $n$ rows, each row $\rho_i$ is generated following the same rule:

$$
\rho_{i,j} = \dfrac{\exp(\nu_{i,j})}{\sum_{j=1}^{n} \exp(\nu_{i,j})},
$$
where $\nu_{i,j}$, ${i,j=1,...,n}$  is i.i.d. sampled by standard Normal distribution $\mathcal{N}(0,1)$.

\subsection{Depth and Width of the Neural Network}
\label{DepWidNN}

In this subsection, we present an experiment on the depth and width of the neural network. Arguably, a wider or deeper neural network gives larger model capacity. However, for the task of choice modeling, we do not recommend a too wide or too deep neural network, mainly for two reasons. First, a complicated neural network architecture requires more number of samples to train, and there might be not enough amount of samples in the application contexts of choice modeling. Second, the customer choice behavior may not need a too complicated neural network to describe.

\begin{figure}[ht!]
    \centering
    \begin{subfigure}[b]{0.4 \textwidth}
        \centering
        \includegraphics[width=1.\textwidth]{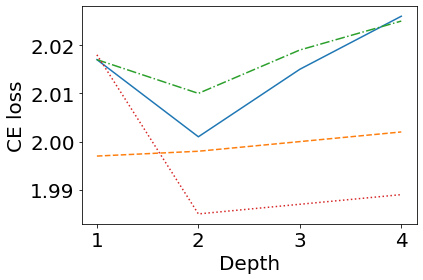}
    \end{subfigure}%
    \begin{subfigure}[b]{0.68 \textwidth}
        \centering
        \includegraphics[width=1.\textwidth]{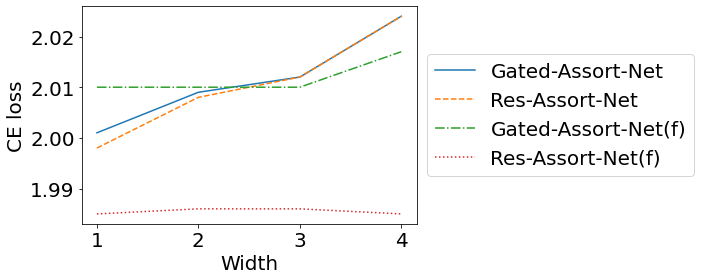}
    \end{subfigure}
    \caption{Network structures and corresponding CE losses.}
    \label{fig:LenWidNN}
\end{figure}

Figure \ref{fig:LenWidNN} presents an experiment on varying the width and depth of the four networks to fit an MCCM with product features. We do not observe a significant change of performance by further widening or deepening the architecture. So, throughout the paper, we mainly set $\text{Depth} = 2$ and $\text{Width}=1$. 

\textbf{Experiment details of Figure \ref{fig:LenWidNN}:}

The underlying model is an MCCM with product features. The sample generation method is the same as that of Table \ref{tabFeat1} and it is deferred to Appendix \ref{sec:apx_fea_model}. We set the total number of products $n=50$, and the length of product features $d=5$. For choice networks with features, we use an encoder containing one middle layer to encode each product feature vector to its latent utility. The product encoder is a fully connected layer $\text{Linear}(5,5)$ followed by another fully connected layer $\text{Linear}(5,1)$. And for choice networks without features, we ignore product features and simply take assortments as input. 

We carry out our training on $100,000$ samples and the testing set has size $10,000$, and we tune the structure of each neural network model to compare their performances on the testing set.
\begin{itemize}
    \item Gated-Assort-Net: $\text{Depth}=1$ indicates a network with no hidden layer, just a fully connected layer $\text{Linear}(50,50)$ that takes as input the assortment and outputs the final layer. $\text{Depth}=2$ indicates a network with one hidden layer, which is a $\text{Linear}(50,50)$ followed by another $\text{Linear}(50,50)$, and $\text{Depth}=3$ indicates two hidden layers. For the experiment that tunes the width, we fix $\text{Depth}=2$, that is one hidden layer. And we tune the size of the hidden layer. $\text{Width}=2$ indicates a $\text{Linear}(50, 100)$ followed by a $\text{Linear}(100,50)$, and $\text{Width}=3$ indicates a $\text{Linear}(50, 150)$ followed by a $\text{Linear}(150,50)$.
    \item Res-Assort-Net: Depth is exactly the number of residual blocks. For the experiment that tunes the width, we fix that there is one residual block, though this block contains one hidden layer. $\text{Width}=2$ indicates that the fully connected part of the residual block is a
    $\text{Linear}(50, 100)$ followed by a $\text{Linear}(100,50)$, and $\text{Width}=3$ indicates a $\text{Linear}(50, 150)$ followed by a $\text{Linear}(150,50)$.
    \item Gated-Assort-Net(f): Following the same rule introduced above, product features are encoded into a vector of latent utilities. The meanings of depth and width are exactly the same as in Gated-Assort-Net.
    \item Res-Assort-Net(f): According to the structure of the Res-Assort-Net with feature, latent utilities and assortments are concatenated into a vector of length $100$, as the input of the residual blocks. Set aside the ending fully connected layer $\text{Linear}(100, 50)$, the rest parts of the residual blocks are exactly the same as ones in Res-Assort-Net. And so depth is the number of residual blocks, and width indicates the size of the hidden layer. $\text{Width}=2$ indicates that the fully connected part of the residual block is a
    $\text{Linear}(100, 200)$ followed by a $\text{Linear}(200,100)$, and $\text{Width}=3$ indicates a $\text{Linear}(100, 300)$ followed by a $\text{Linear}(300,100)$.
\end{itemize}

\subsection{Network Augmentation with Warm Start}

One drawback of the models proposed in this paper is that they all require the set of available products is fixed, i.e., $\mathcal{N}=\{1,...,n\}$ is fixed. When an additional set of products $\mathcal{N}'=\{n+1,...,n'\}$ is available, the model needs to be inevitably retrained on the new product set of $\mathcal{N} \cup \mathcal{N}'$. However, in this subsection, we show that the well-trained network of $\mathcal{N}$ can provide a good warm start for retraining the new network. Specifically, we compare the training procedure of the following two plans:

\begin{itemize}
\item Cold start: We ignore the previous network of $\mathcal{N},$ and train a new network from scratch for the new set $\mathcal{N} \cup \mathcal{N}'$.
\item Warm start: We initialize the weights of the new network according to the previously well-trained network of $\mathcal{N}$.
\end{itemize}

\begin{figure}[ht!]
    \centering
    \begin{subfigure}[b]{0.5 \textwidth}
        \centering
        \includegraphics[width=1.\textwidth]{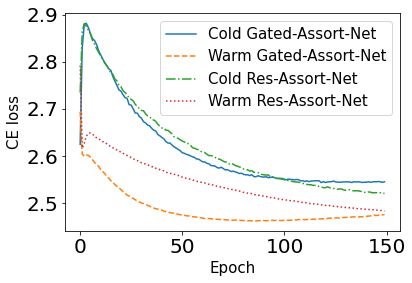}
    \end{subfigure}%
    \begin{subfigure}[b]{0.5 \textwidth}
        \centering
        \includegraphics[width=1.\textwidth]{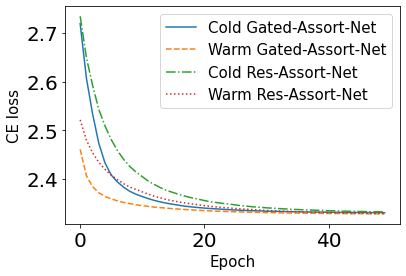}
    \end{subfigure}
    \caption{Validation losses using cold start and warm start: The left panel uses  $2,000$ samples for training and the right panel uses $100,000$ samples for training.}
    \label{fig:WarmStart}
\end{figure}


Figure \ref{fig:WarmStart} presents the validation losses under the two training schemes for the two models with the old set $|\mathcal{N}|=20$ and the new set $|\mathcal{N}'|=5$ . The experiment shows that, when there is an abundant amount of data for augmented models, warm start training will converge to the optimal solution faster in starting stages as the right panel. More importantly, when there is not enough data, warm start training will not only be faster but also produce a better solution. The second observation can be practically useful, as the retails may carry out small-scale experiments to test if it is profitable to introduce more products to their shops. The experiment tells that the neural networks can be trained with a small amount of new data if most of its parameters are inherited from a well-trained model.

\textbf{Experiment details of Figure \ref{fig:WarmStart}.}

We use MCCM to generate the synthetic data for Figure \ref{fig:WarmStart}. We first generate the arriving probability $\lambda$ and the transition matrix $\rho$ for the underlying MCCM. We consider $25$ products and so $\lambda \in \mathbb{R}^{26}, \rho \in \mathbb{R}^{26\times 26}$, as product 0 (the no-purchase option) needs to be included in addition. Following this MCCM we generate $200,000$ pieces of samples, with each sample containing an assortment and the corresponding choice probability and the one-hot encoded final choice vector. We call this synthetic dataset D-augment. Then, based on the $\lambda$ and $\rho$ for D-augment, we can do some modifications to classify product $21$ to $25$ into product $0$. We get $\lambda'\in \mathbb{R}^{21}$ such that $\lambda'_i=\lambda_i$ for $i=1,...,20$ and $\lambda'_0=\lambda_0+\sum_{i=21}^{25}\lambda_i$. In similar manners we get $\rho'\in \mathbb{R}^{21\times 21}$, corresponding to a shrunk version of the original MCCM, i.e., a new MCCM with total number of products $n=20$ plus one no-purchase option. For this new model we also generate $200,000$ pieces of samples, and we call this dataset D-shrink.

We first train models on dataset D-shrink, the Gated-Assort-Net has one hidden layer of size $50$, that is a Linear$(20, 50)$ followed by a Linear$(50, 20)$, and the Res-Assort-Net has one residual block, also with one hidden layer of size $50$. We train the Gated-Assort-Net and Res-Assort-Net on $100,000$ training samples. After that, we consider the augmented dataset D-augment, and we train Gated-Assort-Net and Res-Assort-Net with one hidden layer of size $50$. But we consider two methods for training the augmented model. One is the cold start method, in which we randomly initialize the model parameters and train the model on D-augment; the other is the warm start method, in which we initialize parameters at corresponding positions with the models trained on D-shrink and then train the model on D-augment. The results are given in Figure \ref{fig:WarmStart}, on which we plot how validation loss varies with training epochs. Then left panel is the training on D-augment with training sample size $m=2,000$, and the right panel is the training with sample size $m=100,000$.

\subsection{Tuning Architectures for Neural Network Models for Table \ref{tabRP}}

In this subsection, we describe the experiment setup of Table \ref{tabRP} and extend the results therein with different architectures for the two neural networks. 

\textbf{Experiment setup of Table \ref{tabRP}:}

The synthetic data for different underlying choice models are generated in following ways:
\begin{itemize}
    \item MNL: We sample the mean utility $u_i\in \mathbb{R}$ for each product $i=1,...,n$ i.i.d. from standard Normal distribution  $\mathcal{N}(0,1)$. 
    \item MCCM:  During experiment, we observe that if we generate all rows of the transition matrix $\rho$ in an i.i.d. way, then the choice probability can be well predicted by an MNL model. To distinguish underlying true MCCM from an MNL model, we consider the following generation mechanism. The intuition is that, we assume the set of all products can be divided evenly into several groups, and when a wanted product turns out to be unavailable, the customer is most likely to choose an alternative within the same group of this product. Then it should be unreasonable to assign some certain utility to each product, such that products having larger utilities will always be more likely bought than products having smaller ones.
    
    We considering using $\sigma$ to control deviation ($\sigma = 2.5$ for dataset $n=20$ and $\sigma = 4$ for dataset $n=50$). We also set cluster number $c_{\text{num}}$ ($c_{\text{num}} = 4$ for dataset $n=20$ and $c_{\text{num}} = 10$ for dataset $n=50$) to denote the number of groups. Then arriving probability $\lambda$ is generated by:
    $$
\lambda_i = \dfrac{\exp(\mu_i)}{\sum_{j=1}^{n} \exp(\mu_j)},
$$
where $\mu_i$, $i=1,...,n$ is identically and independently (i.i.d.) sampled by Normal distribution $\mathcal{N}(0,\sigma^2)$. And the transition matrix $\rho$ is generated row by row. For all the $n$ rows, each row $\rho_i$ is generated following:

$$\rho_{i,j}=\dfrac{\exp(\nu_{i,j})}{\sum_{j=1}^{n} \exp(\nu_{i,j})},$$
where $\nu_{i,j}$, $i,j=1,...,n$ is independently sampled by Normal distribution $\mathcal{N}(\bar{\nu}_{i,j},\sigma^2)$ and
\begin{equation*}
\bar{\nu}_{i,j} =
    \begin{cases}
      2\sigma & \forall (i,j)\in \left\{(i,j)|\exists k\in \{0,...,c_{\text{num}}-1\}, kn/c_{\text{num}}< i,j\leq  (k+1)n/c_{\text{num}} \right\} \\
      0  & \text{otherwise}
    \end{cases}.       
\end{equation*}

    Synthetic samples are then generated based on this underlying MCCM. 
    \item NP: the non-parametric (NP) choice model assumes that there is a fixed number of alternating schemes/types amongst customers. Each alternating scheme denotes a permutation of the total set of products, and following this scheme customers will check from the first product to the last until they find a product that is in the assortment. The formal definition is introduced in Section \ref{sec:beyond_rum}. Let $n_{\text{perm}}$ denote the number of candidate permutations with positive probability, we set $n_{\text{perm}} = 10$ for dataset $n=20$ and $n_{\text{perm}}=20$ for dataset $n=50$, and the parameters for the underlying model are the $n_{\text{perm}}$ permutations of $\{1,...,n\}$ and their corresponding probabilities. These parameters are randomly uniformly generated, and following the ground truth non-parametric model we generate synthetic samples.
\end{itemize}

\textbf{More architectures:}

Table \ref{tabRPold} extends Table \ref{tabRP} with more architectures, and it reports the out-of-sample CE loss. We observe that a training set of size $5,000$ is enough for predicting dataset having $20$ products, but is often not enough for predicting dataset with $50$ products. Also, coinciding with our experiments in Appendix \ref{DepWidNN}, we find that in the presence of abundant training samples, the depth and width of the network does not affect much the performance of the model.

In Table \ref{tabRPold}, the row headers represent model types. Gated-Assort-Net$(n, n)$ denotes a Gated-Assort-Net with no hidden layer(the main body of the net is just a Linear$(n, n)$). $n$ denotes the number of total products, which is 20 or 50 in this experiment.  Gated-Assort-Net$(n, 100, n)$ has one hidden layer of size $100$ (the main body is a $Linear(n, 100)$ followed by a Linear$(100, n)$). Res-Assort-Net$(n, n)*2$ is a Res-Assort-Net with 2 residual blocks, and each residual block has no hidden layer(the main body is a Linear$(n, n)$). Res-Assort-Net$(n, n)*3$ is a Res-Assort-Net with 3 residual blocks.

\begin{table*}[ht!]
    \centering
    \begin{tabular}{c|c|cc|cc|cc}
    \toprule
        &\# Samples&MNL20&MNL50  &MCCM20&MCCM50&NP20&NP50 \\ 
         \midrule
   \multirow{3}{*}{Gated-Assort-Net$
(n, n)$}  & 1000&2.00&3.04& 1.97& 0.54& 1.69& 2.71\\
   & 5000& 1.89& 2.81& 1.88&0.48&1.55&2.43\\
   &100000& 1.86& 2.63& 1.84& 0.4&1.50&2.23\\
\midrule
   \multirow{3}{*}{Gated-Assort-Net
$(n, 100, n)$}  & 1000&2.06&3.01& 2.03& 0.67& 2.31& 2.84\\
   & 5000& 1.96&2.85& 1.94& 0.62& 1.66& 2.60\\
   &100000& 1.87&2.66& 1.85&0.43& 1.48& 2.25\\
\midrule
   \multirow{3}{*}{Res-Assort-Net
$(n, n)*2$}  & 1000&2.05&3.05& 2.03& 0.61&1.80&2.79\\
   & 5000&1.93&2.82&1.91&0.56&1.56& 2.51\\
   &100000& 1.86& 2.64& 1.84&0.41&1.49& 2.22\\
\midrule
   \multirow{3}{*}{Res-Assort-Net
$(n, n)*3$}  & 1000&2.08&3.17& 2.04& 0.70& 1.78& 2.90\\
   & 5000&1.94& 3.26& 1.93& 0.55& 1.59& 2.55\\
   &100000&1.86& 2.65& 1.84&0.42&1.48& 2.24\\
   \bottomrule
    \end{tabular}
    \caption{Performance on synthetic data without feature (extended from Table \ref{tabRP}).}
    \label{tabRPold}
\end{table*}

\section{More Details of Numerical Experiments with Features}

\subsection{More Details for Table \ref{tabFeat1}}
\label{sec:apx_fea_model}
\paragraph{Feature-based choice models} \

Given an assortment $\mathcal{S}\subset\mathcal{N}$ and an input feature vector $z_i\in \mathbb{R}^d$ for product $i\in\mathcal{N}$ (which can be product feature along or the concatenation of customer feature and product feature):
\begin{itemize}
    \item \textbf{Feature Based MNL Choice Model:} It assumes the mean utility $u_i(z_i)$ for product $i\in \mathcal{N}$
    $$u_i(z_i)=z_i^\top \beta,$$
    where $\beta \in \mathbb{R}^d$ is the coefficients of features. See \citep{javanmard2020multi} for a pricing problem using the model.
    \item \textbf{Feature Based Markov Chain Choice Model:} This model is from \cite{article}. For product $i\in \mathcal{N}$, the arriving probability $\lambda_i$
      $$\lambda_i(z_i) = \frac{\exp(z_i^\top \beta )}{\sum_{j=1}^n\exp( z_j^\top \beta  )},$$
    where $\beta \in \mathbb{R}^d$. And the vector $\rho_i$ in transition matrix
    $$\rho_{i,j}(z_i) = \frac{\exp(A_jz_{i})}{\sum_{j=1}^n\exp(A_{j}z_i)},$$ 
    where $A\in \mathbb{R}^{n\times d}$ and $A_j$ is $j$-th row of $A$. Then the choice probabilities can be computed  as in MCCM based on $\lambda$ and $\rho$.
\end{itemize}

\paragraph{Training data generation}  \ 

 For the features, we generate the $j$-th feature $z_{i,j}$ of product $i$ from standard Normal distribution $\mathcal{N}(0,1)$ independently (i.e. $z_{i,j}\sim \mathcal{N}(0,1)$ i.i.d.). To model parameters ($A, \beta$), we can generate their entries i.i.d. also from standard Normal distribution $\mathcal{N}(0,1)$. For the arrival probabilities $\lambda_i$'s in feature-based MCCM, we randomly generate $\lambda$ in the same way as the feature-free case in Appendix \ref{sec:apx_ass_eff_A1}.

\subsection{More architecture tuning}\

\begin{table*}[ht!]
\centering
\begin{tabular}{c|c|c|c|c}

\midrule
 MCCM &Gated configuration & CE loss  & Res configuration & CE loss\\
\midrule
\multirow{2}{*}{Benchmark}
& Oracle & 1.932& Oracle & 1.932 \\
& MNL-MLE(f) & 2.212& MNL-MLE(f) & 2.212 \\
\midrule
\multirow{4}{*}{Assort-Net}
& [50,50] & 2.010 & [50,50] & 1.988\\
& [50] & 2.017 & [50] & 1.997\\
& [50,50,50] & 2.015 & [50,50,50] & 2.000\\
& [100,50] & 2.009 & [100, 50] & 2.008\\
\midrule
\multirow{10}{*}{Assort-Net(f)}
& [[1], [50,50]] & 2.010 & [[1], [100,50]] & 1.984\\
& [[5,1], [50,50]] & 2.010 & [[5,1], [100,50]] & 1.985\\
& [[5,5,1], [50,50]] & 2.008 & [[5,5,1], [100,50]] & 1.985\\
& [[5,5,5,1], [50,50]] & 2.007 & [[5,5,5,1], [100,50]] & 1.984\\
& [[10,1], [50,50]] & 2.008 & [[10,1], [100,50]] & 1.984 \\
& [[15,1], [50,50]] & 2.009 & [[15,1], [100,50]] & 1.986\\
& [[20,1], [50,50]] & 2.008 & [[20,1], [100,50]] & 1.983\\
& [[5,1], [50]] & 2.017 & [[5,1], [50]] & 2.018\\
& [[5,1], [50,50,50]] & 2.019 & [[5,1], [100,100,50]] & 1.987\\
& [[5,1], [100,50]] & 2.010 & [[5,1], [100,50]] & 1.985\\

\bottomrule
    \end{tabular}
    \caption{Extension of Table \ref{tabFeat1} with more architectures. The data is generated by feature-based MCCM.}
    \label{tabFeat}
\end{table*}


Table \ref{tabFeat} extends Table \ref{tabFeat1} with more neural network architectures. We find the model performance is quite stable with respect to architecture tuning. 


For neural networks without features (under ``Assort-Net''), the model configurations for Gated-Assort-Net, $[50]$ denotes a one layer Linear$(50, 50)$, and $[50, 50]$ denotes a two-layer network, Linear$(50, 50)$ followed by a Linear$(50, 50)$. $[50, 50, 50]$ denotes a three-layer network, and $[100, 50]$ denotes a two-layer, Linear$(50, 100)$ followed by a Linear$(100, 50)$. For the Res-Assort-Net part, $[50]$ denotes one residual block, $[50, 50]$ denotes two residual blocks, $[50, 50, 50]$ three residual blocks, and $[100, 50]$ denotes a Res-Assort-Net with one residual block but contains a hidden layer of size $100$.

For neural networks with features (under ``Assort-Net(f)''),  in the ``Gated configuration'' and ``Res configuration'' columns, the first list ($[1]$ or $[5,1]$ or $[5,5,1]$ etc.) denotes the structure of the product encoder, and the second list ($[50]$ or $[50,50]$ etc.) denotes the structure of the assortment network. The meanings of the configuration lists for the assortment network are generally the same as the feature-free cases, though some additional explanations need to be devoted to the Res-Assort-Net(f). According the network structure, the assortment network takes as input the concatenated vector of latent utilities and assortments, leading to a vector of length $100$. $[100, 50]$ indicates one residual block with the linear part Linear$(100, 100)$ and an ending layer Linear$(100, 50)$. $[50]$ indicates no residual block and just an ending layer Linear$(100, 50)$. $[100, 100, 50]$ indicates two residual blocks with linear part Linear$(100, 100)$ and an ending layer Linear$(100, 50)$. The first list gives the structure of product encoder. Recall that the length of the product feature $d=5$, and product features are encoded into latent utility values by several layers of fully connected network. $[1]$ denotes a single layer Linear$(5,1)$, $[5,1]$ denotes a Linear$(5,5)$ followed by a Linear$(5,1)$. $[10,1]$ denotes a Linear$(5,10)$ followed by a Linear$(10,1)$. The rests follow similar rules.


\subsection{More Details on Real Data Experiments}
\label{sec: real_data_exp}
\paragraph{Data Description}
\begin{itemize}
    \item \textbf{SwissMetro}   The SwissMetro Data Set is a public survey data set collected to analyze preference between the Metro system against usual transport modes including car and train.  The description of the dataset is given in \cite{SwissMetro}. For convenience of network training, we clean the data by removing samples whose features are taking abnormal values or values too rarely occur. To be specific, we only keep samples whose ``CHOICE'' (final choice) is not $0$ (which indicates unknown), ``WHO'' (who pays the ticket) value is not $0$ (which indicates unknown), ``AGE'' is not $6$ (which indicates unknown), ``INCOME'' is not $4$ (which indicates unknown), and ``PURPOSE'' is in $[1,2,3,4]$ (1: Commuter, 2: Shopping, 3: Business, 4: Leisure). After cleaning, there are in total $9,135$ samples and the number of alternatives in assortment varies from $2$ to $3$: only when owning the car the customer can have that choice. The sample sizes in train, validate and test process are $7,000$, $1,000$ and $1,000$ correspondingly. There are  $8$ customer features and $3$ product features, shown in Table \ref{tab:Features_real}. For product features. Note that ``CAR'' has no ``HE'' (Headaway, which means interval time between two consecutive train/metro/car arrivals) feature in the original dataset, so we set it to $0$. This is reasonable since for private cars the waiting time is always $0$. In this dataset the product features are not static, as they vary among different types of customers.

\item \textbf{Expedia Search} The Expedia Data Set is a public data set from Kaggle competition ``Personalize Expedia Hotel Searches - ICDM2013''. It contains the ordered list of hotels according to the user's search. We clean the data by dropping out the searches with outlier prices ($>\$1,000$) and days of booking in advance (gap between booking date and check-in date $>365$) and also the features with missing values. After cleaning, there are in total $386,557$ searches where the size of the offered hotel/searched results list (i.e., the number of alternatives in one assortment) is between $5$ to $38$, and $136,886$ unique hotels. The sample sizes in train, validate and test process are $200,000$, $20,000$ and $20,000$ correspondingly. There are  $6$ customer features and $7$ product features as shown in Table \ref{tab:Features_real}.

For assortment vector $S$, we encode the ranked positions in assortments ($1-39$ with $39$ as no-purchase option) and if some position has no shown product, we assign their product features as $0$. Thus the assortment vector captures the ranking and assortment structure within the assortment. We should notice we only use the positions of products in offered assortments instead of the products themselves. There are several reasons: The number of unique products is too large (recall there are $136,886$ unique hotels in $386,557$ offered assortments after cleaning) to be covered as assortment vector; Also in practice, hotels with different locations should never be included in the same assortment and the assortment effect from such hotels should be ignored; Further, $39$, the largest size of assortment, should be large enough to capture all possible hotels customers are interested in and we can further utilize the product features to identify hotels with the same ranked positions in different assortments.
\end{itemize}
\begin{table}[ht!]
    \centering
    \begin{tabular}{c|c|c}
    \toprule
           Data&Customer Feature&Product Feature \\
\midrule
 \multirow{8}{*}{SwissMetro} 
 &Gender& Travel time\\
 &Age&Headaway\\
 &Income&Cost\\
 &First class or not&\\
 &Who pays (self, employer...)&\\
 &Purpose (commuter, shopping...)&\\
 &\# luggages&\\
 &Owning annual ticket or not&\\
 \midrule
 \multirow{7}{*}{Expedia} &\# nights stay &Hotel star rating\\
 &\# days booking in advance &Hotel chain or not\\
 &\# adults& Location score (by Expedia)\\
 &\# children& Historical prices\\
 &\# rooms &Price\\
 & Saturday included or not&Promotion or not\\
 &&Randomly ranked or not (by Expedia)\\
 \bottomrule
    \end{tabular}
    \caption{Features in SwissMetro and Expedia.}
    \label{tab:Features_real}
\end{table}

\paragraph{Tuning Architectures for Neural Networks and Performances}

The performances of different models on two datasets are summarized in Table \ref{tab:RD}, which is extended from Table \ref{tabExp}. The configurations of both our assort-nets and the benchmark nets are given in the form of channel arrays. For our assort-nets, the first array denotes the structure of customer encoder, the second array denotes the product encoder, and the third denotes the assort-net part of the whole network. Take the last line of the gated-assort-net for example, in the configuration list $[[30,10],[10,10],[30,30,3/39]]$, the first array $[30, 10]$ denotes the structure of customer encoder, having one hidden layer of size $30$ and a final layer of size $10$. The second array $[10, 10]$ denotes the structure of the product encoder, having one hidden layer of size $10$ and a final layer of size $10$. We guarantee that the size of the final layer of the customer encoder must equal that of the product encoder. The third array $[30, 30, 3/39]$ denotes that the assort-net part has two hidden layers of size $30$. The size of the final layer is dependent on the number of total products $n$, which is $3$ for SwissMetro and $39$ for Expedia. The Res-Assort-Net follows the same way to denote the configurations of customer encoder and product encoder. To denote the configuration of the Res-Assort-Net part, take the last configuration $[6, 60, 6]*2$ for example, it denotes that there are 2 residual blocks, each containing a hidden layer of size $60$.

We defer the description of benchmark algorithms in Appendix \ref{sec:bench_alg}. The configurations ($[100, 9]$ or $[100, 30]$) denote the corresponding network structure, having a hidden layer of size $100$ and a final layer of size $9$ or $30$.

For Random Forest, the hyper-parameters including number of trees and maximum depth are chosen by validation dataset performance. Specifically, the number of trees is tested among $10,20,50,100,200,300,400$ and the maximum depth is tested among $4,6,8,12,16,18,20$. Other parameters are set as default values in the Python package scikit-learn's function \textit{sklearn.ensemble.RandomForestClassifier}. We choose the testing structure of our networks following the rule of not getting over complex. We restrict all three parts of our network (product encoder, customer encoder and assort-net) to a depth of no more than $2$. The widths are tuned a little bit loosely (due to their relative lesser effect on the model performance), but no more than $200$. The architecture of TasteNet follows the original paper \cite{han2020neural}, in which they used 110 hidden units. The architecture of DeepMNL is set to keep some degree of similarity with the architecture of TasteNet.

\begin{table*}[ht!]
    \centering
    \begin{tabular}{cc|c|c|c|c}
    \toprule
         \multicolumn{2}{c}{Model}&\multicolumn{2}{|c}{SwissMetro}&\multicolumn{2}{|c}{Expedia} \\ 
         \midrule
  && CE loss  & Acc. & CE loss&Acc.\\
\midrule
\multirow{6}{*}{Gated-Assort-Net(f)} 
&[[10],[10],[3/39]]&0.696&0.708&2.479&0.300\\
&[[100,10],[10],[3/39]]&0.638&0.733&2.468&0.303\\
&[[10],[10,10],[3/39]]&0.673&0.704&2.458&0.306\\
&[[10],[10],[10,3/39]]&0.708&0.692&2.463&0.306\\
&[[100,10],[10,10],[10,3/39]]&\textbf{0.598}&\textbf{0.759}&2.458&0.306\\
&[[30,10],[10,10],[30,30,3/39]]&0.623&0.734&2.403&0.332\\
\midrule
\multirow{3}{*}{Res-Assort-Net(f)}&
[[100, 10],
[10, 10],
[6, 6]*1]&0.607&0.738&2.468&0.307\\
&
[[100, 10],
[10, 10],
[6,60,6]*1]&0.608&0.736&2.355&0.346\\
&[[200, 20],
[20, 20],
[6,60,6]*2]&0.610&0.758&\textbf{2.325}&\textbf{0.355}\\
\midrule
MNL-MLE& & 0.883&0.581
& 2.827  &0.298\\
\midrule
MNL(f)-MLE& &0.810

&0.621&2.407& 0.331\\
\midrule
\multirow{2}{*}{TasteNet}  
&[100, 9]
 +MNL
&0.691&0.701&--&--\\ 
&[100, 9]
 +MNL(f)
&0.681&0.698&2.436&0.332\\
\midrule
\multirow{2}{*}{DeepMNL} &[100, 30]
 +MNL
&0.751&0.658&--&--\\ 
&[100, 30]
 +MNL(f)
&0.741&0.655&2.374&0.344\\
\midrule
Random Forest &&0.633&0.735&2.458&0.317\\
   \bottomrule
    \end{tabular}
    \caption{Performance on the SwissMetro and Expedia (extended from Table \ref{tabExp}).}
    \label{tab:RD}
\end{table*}

\paragraph{Dealing with Missing Products for TasteNet and DeepMNL}

Due to the fixed size of input vectors when training the neural networks, the DeepMNL and TasteNet cannot be applied directly when the size of assortment will change across different samples. For SwissMetro dataset, TasteNet and DeepMNL are only trained based on the samples with full assortments where all products are offered. The number of such samples in the dataset is $7,839$. We predict the full assortments choice probabilities by using trained TasteNet or DeepMNL, and use trained MNL-MLE or MNL(f)-MLE to predict the other assortments (with missing product) where the sample size is $1,296$ in the dataset. We denote the corresponding prediction models as TasteNet(DeepMNL)+MNL and TasteNet(DeepMNL)+MNL(f); For Expedia dataset, since there are only $9$ out of $386,557$ samples are full assortments, we cannot use the same method as in SwissMetro due to the lacking data (otherwise we are almost reproducing the MNL-MLE or MNL(f)-MLE). Thus, we just assign $0$ into the missing entries of product feature vectors, which is same as Gated-Assort-Net(f) and Res-Assort-Net(f). Since these methods are recovering the missing feature vectors by $0$ and the last layer is softmax, we also name them as TasteNet(DeepMNL)+MNL(f).

\subsection{Benchmark Algorithms}
\label{sec:bench_alg}
\begin{itemize}
 \item MCCM-EM: We follow the work \citep{csimcsek2018expectation} to build the expectation-maximization algorithm based on MCCM. In summary, the algorithm starts with some random initial transition matrix $\rho$ and arrival probabilities $\lambda$. For the expectation step, and based on the current transition matrix and arrival probabilities, the algorithm computes the expected probability of being interested in each product when a customer arrives and the number of times that a customer transitions from one product to another during the decision process for each sample. For the maximization step, it maximizes the likelihood to update the transition matrix and arrival probabilities by treating previous two expected values as true values. The algorithm repeats above two steps until converges. In our experiments, the initial transition matrix and arrival probabilities are chosen by uniformly randomizing each entry and normalizing them to stochastic matrix or distribution vector. We terminate the algorithm when the mean changes of updating transition matrix and arrival probabilities is smaller than $0.01\%$.
 
\item MNL-MLE: We assume the underlying choice model is the MNL model and ignore all observed features (if any). We estimate the unknown mean utilities of products by MLE.

\item MNL(f)-MLE: We assume the underlying choice model is featured based MNL choice model introduced in Appendix \ref{sec:apx_fea_model}, where the feature vector $z_i=(g,f_i)$ is concatenation of customer feature $g$ and product feature $f_i$. We estimate the unknown parameters by MLE.
    
\item TasteNet: We follow the architecture introduced by \cite{han2020neural}. We train a neural network $g_{\theta}(\cdot)$ to encode the customer feature vector $g\in \mathbb{R}^{d'}$ into a vector $(\alpha_1,...,\alpha_n) \coloneqq g_{\theta}(g)\in \mathbb{R}^{nd}$, where $\alpha_i$ is interpreted as the coefficient vector of product features $f_i$. For product $i\in \{1,..., n\}$, its mean utility is encoded by:
    $$
    u_i:=\alpha^\top_if_i.
    $$
    This utility vector $u\in \mathbb{R}^n$ is then passed to a softmax layer, giving the predicted choice probability.
    
    \item DeepMNL: We modify the architecture introduced by \cite{sifringer2020enhancing}.
    We train a neural network $g_{\theta}(\cdot)$ to encode the concatenated feature vector $(g,f_i)\in \mathbb{R}^{d+d'} $of customer features $g\in \mathbb{R}^{d'}$ and product features $f_i \in \mathbb{R}^{d}$ into scalar $g_{\theta}\left( (g,f_i) \right)\in \mathbb{R}$ as the mean utility of product $i$ in MNL model. This utility vector is then passed to a softmax layer, giving the predicted choice probability.
    \item Random Forest: We follow the work \cite{chen2021estimating} to apply the random forest method to predict customer choice. Specifically, for each sample assortment, we concatenate the assortment vector $S$, customer feature $g$ and all product features $f_i$, $i=1,...,n$, as its input (thus same as Gated-Assort-Net(f) and Res-Assort-Net(f)). For the products not offered in assortment, we assign their product feature values as $0$. 
\end{itemize}

\section{Interpreting Neural Networks}

\subsection{Effects of Fully Connected Layers}
\begin{figure}[ht!]
    \centering
    \begin{subfigure}[b]{0.5 \textwidth}
        \centering
        \includegraphics[width=1\textwidth]{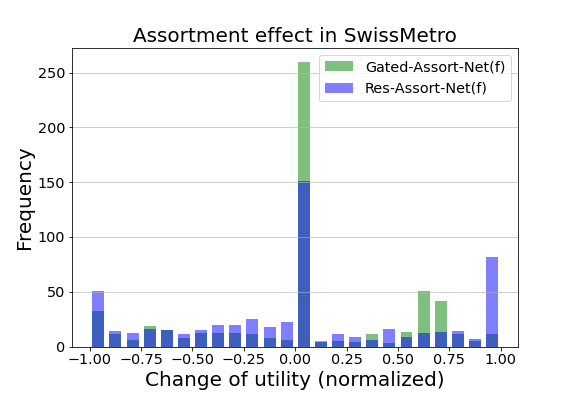}
    \end{subfigure}%
    \begin{subfigure}[b]{0.5 \textwidth}
        \centering
        \includegraphics[width=1\textwidth]{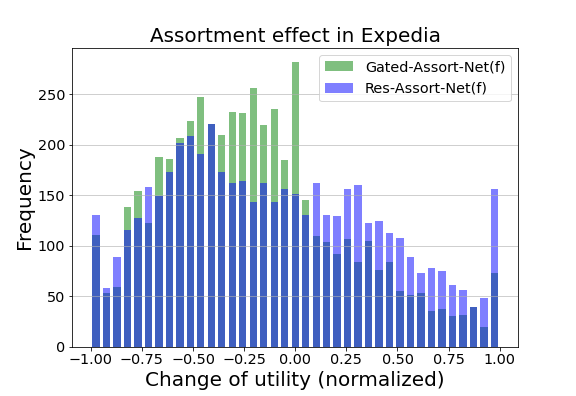}
    \end{subfigure}
    \caption{Effects of fully connected layers in SiwssMetro and Expedia. A larger absolute value means a larger assortment effect from fully connected layers.}
    \label{fig:Eff_layer}
\end{figure}

In this subsection we interpret how the fully connected layers process the latent utilities with assortments information through logic gates in Gated-Assort-Net(f) or directly adding it as input features in Res-Assort-Net(f). Specifically, we compute a normalized difference between the input latent utility and output utility of each network. Such a difference reflects the assortment effect in the network.  We denote such difference as $\Delta u$ and plot the histograms of Gated-Assort-Net and Res-Assort-Net in Figure \ref{fig:Eff_layer}.

From Figure \ref{fig:Eff_layer}, we can find effects from fully connected layers will vary due to different dataset structures and networks.  For the SwissMetro data set, the majority of changes in normalized utilities $\Delta u$ are located in $0$ for both two networks. One potential reason is that there are only three products and only small part of samples ($13.5\%$)  do not offer all products. Thus, the effect of assortments through fully connected layer is very small. However, since the Expedia data has more complicated assortment patterns, the $\Delta u$'s do not concentrate in $0$ anymore and has more variances. This also explains why the Res-Assort-Net(f) that has a stronger involvement of the assortment vector, shows an advantage of the Expedia dataset.

Further, by comparing Gated-Assort-Net(f) and Res-Assort-Net(f) in both data sets, there are two interesting observations: (1) There are more $\pm1$'s in Res-Assort-Net(f), which means more dramatically changes in normalized utilities after passing fully connected layers; (2) The histograms of Res-Assort-Net(f) are more flatten than Gated-Assort-Net(f), which means the influences of fully connected layers have more variations. Both of the above observations can be partially explained by the designed architectures: the Gated-Assort-Net(f) only uses assortments as logical gates but Res-Assort-Net(f) also concatenates it as input features and can squeeze more information to process the input latent utilities, which leads to potentially larger degree of influence with more variations.

\textbf{Experiment details of Figure \ref{fig:Eff_layer}:}

Fixed one assortment $\mathcal{S}$, we omit the product not offered. For each offered product $i\in \mathcal{S}$, we first normalize its input and output utilities of the fully connected layers (from Gated-Assort-Net(f) or Res-Assort-Net(f)) by 
$$\tilde{u}_i^{I/O}=\frac{u^{I/O}_i-\min_{j\in \mathcal{S}}u^{I/O}_j}{\max_{j\in \mathcal{S}}u^{I/O}_j-\min_{j\in \mathcal{S}}u^{I/O}_j},$$
where $u^{I}$ denotes input utilities and $u^{O}$ denotes outputs of the fully connected layers. Note that the normalization is within assortment: different assortments may have different scales on utilities and what matters is the relative value. 

Then we compute the differences of them:
$$\Delta u_i\coloneqq \tilde{u}_i^{O}-\tilde{u}^{I}_i$$
for each $i\in \mathcal{S}$ of the fully connected layers. $\Delta u_i=0$ means the normalized utilities are same before and after the processing of fully connected layers. If $\Delta u_i$ is $-1$ or $1$, product $i$ is dramatically influenced: before processing it has the largest utility (smallest utility), i.e. $\tilde{u}^{I}_i=1$ ($\tilde{u}^{I}_i=0$), but after processing it becomes the smallest (largest). 

We randomly choose $200$ samples from the test data to plot Figure \ref{fig:Eff_layer}. The configurations of Gated-Assort-Net(f) and Res-Assort-Net(f) are chosen as the one with the best performance shown in Table \ref{tab:RD}.

\subsection{More on Model Calibration}

As mentioned earlier, an alternative metric for probabilistic prediction is the calibration performance. Table \ref{tab:Calib} summarizes the calibration performances of different models averaged for all choices. In general, both Gated-Assort-Net(f) and Res-Assort-Net(f) have good performances in calibration, which complements the low out-of-sample CE losses shown in Table \ref{tab:RD} in the sense of reliability.

\begin{table}[ht!]
    \centering
    \begin{tabular}{c|cc}
    \toprule
  \multirow{2}{*}{Model}    &\multicolumn{2}{c}{Datasets}\\
& SwissMetro & Expedia \\
\midrule
MNL-MLE
&4.56&0.79\\
MNL(f)-MLE
&6.22&0.61\\
\midrule
TasteNet+MNL
&5.50&--\\
TasteNet+MNL(f)
&4.66&0.60\\
DeepMNL+MNL
&8.19&--\\
DeepMNL+MNL(f)
&7.26&0.59\\
Random Forest
&7.56&0.41\\
\midrule
Gated-Assort-Net(f)
&4.67&0.39\\
Res-Assort-Net(f)&4.65&0.42\\
 \bottomrule
    \end{tabular}
    \caption{Calibration errors on SwissMetro and Expedia by ACE ($\times10^{-2}$).}
    \label{tab:Calib}
\end{table}

\textbf{Metric calculation for Table \ref{tab:Calib}:}

We use the Adaptive Expected Calibration Error (ACE) as our metric to measure calibration performance \citep{nixon2019measuring}:
$$\text{ACE}=\frac{1}{mB}\sum_{i=1}^n n_i\sum_{b=1}^B|\text{acc}(b,i)-\text{conf}(b,i)|,$$
where $m,B,n$ are the number of samples (assortments), bins and products. Here $n_i$ is the number of assortment samples where product $i$ is included, $\text{acc}(b,i)$, $\text{conf}(b,i)$ are the mean of empirical purchase probabilities (confidence) and mean of predictions (accuracy) in $b$-th bin for product $i$. Further, for each product $i$, the bins are chosen by equal-mass, i.e., each bin will contain the same number of samples. 

For Table \ref{tab:Calib}, we choose the number of bins $B=25$. The configurations of models are chosen as the one with lowest CE losses in Table \ref{tab:RD}. The training and testing data are same as in Appendix \ref{sec: real_data_exp}.

\section{Numerical Experiments on Feature-free Real Data}

Lastly, we conduct one more numerical experiment on a real data that has no product or customer feature. The summary statistics of the dataset is given in Table \ref{tab:Hotel_data}. 

\begin{table}[ht!]
    \centering
    \begin{tabular}{c|c|c|c|c|c}
    \toprule
       &Hotel $1$&Hotel $2$&Hotel$3$&Hotel $4$&Hotel$5$ \\
\midrule
\# Samples&8935&870&6180&1330&1065\\
\# Products&12&7&17&8&7\\
\# Train Samples&6000&600&4000&900&800\\
\# Validate Samples&1000&100&1000&100&100\\
\# Test Samples&1000&100&1000&200&100\\
 \bottomrule
    \end{tabular}
    \caption{Description of Hotel data (\cite{bodea2009data}).}
    \label{tab:Hotel_data}
\end{table}

\begin{table}[ht!]
    \centering
    \begin{tabular}{c|ccccc}
    \toprule
  \multirow{2}{*}{Model}      &\multicolumn{5}{c}{Dataset} \\
       &Hotel $1$&Hotel $2$&Hotel$3$&Hotel $4$&Hotel$5$ \\
\midrule
MNL-MLE
&0.775&0.902&0.980&0.661&0.660\\
MCCM-EM
&0.757&\textbf{0.856}&0.945&\textbf{0.656}&\textbf{0.655}\\

\midrule
Gated-Assort-Net & \textbf{0.748} & 0.869 & 0.934 & 0.678 & 0.680\\
Res-Assort-Net & \textbf{0.748} & 0.879 & \textbf{0.928} & 0.688 & 0.679\\
 \bottomrule
    \end{tabular}
    \caption{Out-of-sample CE loss.}
    \label{tab:Hotel_result}
\end{table}

\paragraph{Data description}
The data  is from \cite{bodea2009data} and it is originally collected from five U.S. properties of a major hotel chain. For each customer, the offered products in assortment are the different room types, such as king room smoking and 2 double beds room non-smoking. We preprocess the data similar as \cite{csimcsek2018expectation}, including removing the purchases without matched room type in assortment, the room type with rare purchases, and adding the fake no purchase result. 

Table \ref{tab:Hotel_result} demonstrates the performance of two benchmark methods and the two neural network models. In this experiment, we train our Gated-Assort-Net and Res-Assort-Net for each of hotels using only samples for the hotel. Both networks are taking the simplest configuration: Gated-Assort-Net has no middle layer and Res-Assort-Net has one residual block. Compared to the benchmark MCCM-EM algorithm, we observe that our models have better performance on the dataset of Hotel $1$ and Hotel $3$. This is due to the fact that these two datasets have much larger number of samples than the other $3$ hotels. This coincides with our previous observations that,  a moderately large number of samples is necessary to fully show the advantage of a neural network model. When the number of samples is small, or possibly, the customer choice pattern is simple, the more analytical model of MNL-MLE or MCCM-EM is preferrable.

\end{document}